\documentclass{article}
\usepackage[nonatbib,preprint]{neurips_2025}

\usepackage[utf8]{inputenc} 
\usepackage[T1]{fontenc}    
\usepackage{microtype}      

\usepackage{bm,bbm}
\usepackage{amsfonts,amsmath,amssymb,amsfonts,amsthm}
\usepackage{mathtools,nicefrac}

\usepackage{hyperref,url} 
\usepackage{cleveref}

\usepackage{textcomp}
\usepackage{xcolor}

\usepackage{comment}
\usepackage{titlesec,enumitem}
\usepackage{flushend}

\usepackage{graphicx,wrapfig}
\usepackage{booktabs,colortbl,multirow,multicol}

\usepackage[numbers]{natbib}
\usepackage{adjustbox}

\usepackage[linesnumbered,ruled,vlined]{algorithm2e}
\usepackage{amsmath,amssymb}
\usepackage{graphicx}
\usepackage[font=small]{caption}
\usepackage{cleveref}

\Crefname{equation}{Eqn.}{Eqns.}

\theoremstyle{plain}
\newtheorem{theorem}{Theorem}[section]
\newtheorem{proposition}[theorem]{Proposition}
\newtheorem{lemma}[theorem]{Lemma}
\newtheorem{corollary}[theorem]{Corollary}
\theoremstyle{definition}

\theoremstyle{remark}

\newcommand{\modelname}{DYMAG}

\newcommand{\chaosmath}{$u_S(v,t)$}

\newcommand{\chaoseqname}{Sprott}

\newcommand{\githubanon}{\url{https://github.com/KrishnaswamyLab/DYMAG}}

\newcommand{\best}[1]{\textbf{\textcolor{blue}{#1}}}
\newcommand{\second}[1]{\textbf{\textcolor{purple}{#1}}}

\newcommand{\bestmm}[1]{\textcolor{blue}{\mathbf{#1}}}
\newcommand{\secondmm}[1]{\textcolor{purple}{\mathbf{#1}}}

\begin{document}

\title{DYMAG: Rethinking Message Passing Using Dynamical-systems-based Waveforms}

\author{
\textbf{Dhananjay Bhaskar}$^{1,*}$ \quad
\textbf{Xingzhi Sun}$^{1,*}$ \quad
\textbf{Yanlei Zhang}$^{2,*}$ \quad
\textbf{Charles Xu}$^{3,*}$ \\
\textbf{Arman Afrasiyabi}$^{1}$ \quad
\textbf{Siddharth Viswanath}$^{1}$ \quad
\textbf{Oluwadamilola Fasina}$^{1}$ \\
\textbf{Maximilian Nickel}$^{5}$ \quad
\textbf{Guy Wolf}$^{2,4}$ \quad
\textbf{Michael Perlmutter}$^{6}$ \quad
\textbf{Smita Krishnaswamy}$^{1,\dagger}$ \\
\\
$^1$Yale University \quad
$^2$Mila – Quebec AI Institute \quad
$^3$MIT \\
$^4$Université de Montréal \quad
$^5$Meta \quad
$^6$Boise State University \\
$^*$Equal contribution \quad
$^\dagger$Corresponding author: \texttt{smita.krishnaswamy@yale.edu}
}
\maketitle

\begin{abstract}
We present DYMAG, a graph neural network based on a novel form of message aggregation. Standard message-passing neural networks, which often aggregate local neighbors via mean-aggregation, can be regarded as convolving with a simple rectangular waveform which is non-zero only on 1-hop neighbors of every vertex. 
%
Here, we go beyond such local averaging. We will convolve the node features with more sophisticated waveforms generated using 
dynamics such as the heat equation, wave equation, and the Sprott model (an example of chaotic dynamics). 
 Furthermore, 
we use snapshots of these dynamics at different time points to create waveforms at many effective scales. Theoretically, we show that these dynamic waveforms can capture salient information about the graph including connected components, connectivity, and cycle structures even with no features. Empirically, we test DYMAG on both real and synthetic benchmarks to establish that DYMAG outperforms baseline models on recovery of graph persistence, generating parameters of random graphs, as well as property prediction for proteins, molecules and materials. 
Our code is available at \githubanon.
\end{abstract}
\section{Introduction}
\label{sec:intro}

Message passing graph neural networks (GNNs) rely on aggregating signals via local averaging, which can be interpreted as convolving the node features with a simple, rectangular waveform that is non-zero only within one-hop neighborhoods of each vertex. It is known that this type of message-passing tends to suffer from over-smoothing if too many iterations are applied and from under-reaching if too few are applied~\citep{rusch2023survey,keriven2022not,errica2023adaptive}. One possible solution is to use multiscale message passing 
\citep{abu2019mixhop}. Another approach, \cite{gao2019geometric,perlmutter2023understanding,gama2018diffusion,gama2019stability,jiang2024limiting,wenkel2022overcoming,bodmann2024scattering,xu2023blis,tong2022learnable} more directly related to our work, is to use graph wavelets \cite{hammond2011wavelets, coifman2006}. 
These wavelets can be viewed as convolving the input features with multiscale, oscillatory waveforms, in contrast to the simple, rectangular, one-hop waveforms used in message passing.



Here, we introduce DYMAG which uses dynamics on the graph to generate waveforms which we will convolve with the node features. 
We will use these waveforms as a form of multiscale
 message aggregation, which we show can effectively extract graph geometric and topological information and outperform baseline methods on graph-level tasks that rely on such graph properties. 

We evaluate DYMAG on a broad spectrum of graph learning benchmarks spanning synthetic, citation, molecular, and materials science datasets. To assess its ability to recover generative and topological structure, we first test on synthetic graphs, including Erd\H{o}s-R\'enyi and stochastic block models, where the task involves inferring graph parameters and persistent features. We then evaluate on citation networks, including homophilic datasets - Cora~\citep{mccallum_automating_2000}, Citeseer~\citep{giles_citeseer_1998}, and PubMed~\citep{sen_collective_2008} - and heterophilic datasets - Texas, Wisconsin, and Cornell~\citep{pei2020geom}. We further demonstrate DYMAG’s scalability on the largest dataset in the Open Graph Benchmark, \texttt{ogbn-papers100M}~\citep{hu2020open}, demonstrating that it can recover topological properties of massive graphs. For molecular property prediction, we consider both protein graphs PROTEINS, ENZYMES, and MUTAG \citep{dobson_distinguishing_2003} and small-molecule graphs (DrugBank~\citep{wishart_drugbank_2018}, Drug Therapeutics Program AIDS Antiviral Screen Data~\citep{nih_aids_screen}). 
Finally, we test on the Materials Project dataset~\citep{10.1063/1.4812323} to predict materials properties such as band gaps. Across these varied domains, DYMAG consistently outperforms standard GNNs and approaches the performance of pretrained, domain-specific models.
Our main contributions are as follows: 

\noindent\begin{minipage}{\linewidth}
\begin{enumerate}[leftmargin=1.5em, itemsep=0pt, topsep=0pt]
\item We introduce a new type of GNN called DYMAG, which uses dynamics-waveform-based message aggregation, and is capable of capturing complex signal patterns on a graph.
\item We show theoretically that our waveforms 
capture both the low-pass and band-pass portion of the input features as well as geometric and topological information including the graph spectrum, connected components, connectivity, cycles, shortest-path distance, and curvature. 
\item We show that our method better predicts geometric and topological network properties—such as curvature and extended persistence images—compared to standard message passing networks.
\item We demonstrate that \modelname{} outperforms various message passing networks 
as well as large pretrained domain-specific model on molecular predictions.
\end{enumerate}
\end{minipage}

\begin{figure} [!htbp]
\centering
\scalebox{0.9}{
    \begin{minipage}{0.55\textwidth}
    \centering
    {\scriptsize (a) Waveforms of Message Passing vs Diffusion Wavelets and DYMAG}
    \includegraphics[width=\linewidth]{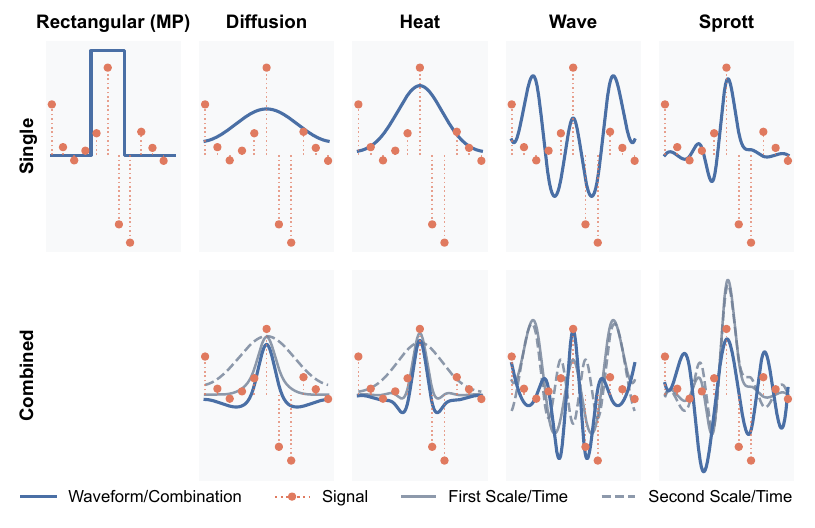}
    \end{minipage}
    \hfill
    \begin{minipage}{0.44\textwidth}
    \centering
   {\scriptsize (b) frequency domain of waveforms and combinations }
    \includegraphics[width=\linewidth]{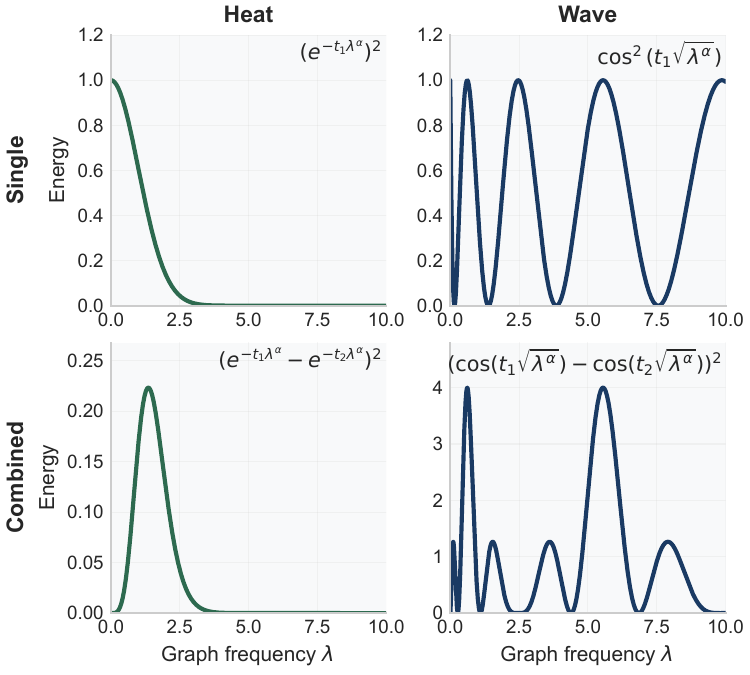}
    \end{minipage}
    }
\caption{\textbf{Visualization of Waveforms} (a) Waveforms visualized on a line graph with a signal (feature), where DYMAG provides more diverse waveforms than standard message passing; (b) waveforms and combinations provide low-pass and bandpass filters in the frequency domain.
}\label{fig:waveforms}
\end{figure}
\vspace{-10pt}

\section{Related work and Background}\label{sec:bg}
Previous work has either analyzed dynamics on graphs \citep{reijneveld2007application,boccaletti2014structure,simoes2011agent,holme2012temporal,belbute2020combining,sanchez2020learning,pfaff2020learning} or aimed to use dynamics as a framework for understanding GNNs. In the  latter case \citet{chamberlain2021beltrami,chamberlain2021grand,eliasof2022} and \citet{thorpe2022grand++}  viewed message passing as a time-discretized diffusion PDE and used this insight to design novel GNNs. 
Unlike those methods, we view PDE solutions as waveforms and use convolution against these waveforms to define our aggregation rule. Additionally, existing work primarily focus on parabolic equations while we also consider hyperbolic and chaotic dynamics. 
We provide a further discussion of related work in  \Cref{sec: related}.
Now, we provide backgrounds on graph signal processing and dynamics. See \Cref{appx:sec:bg} for details.

\subsection{Graph Signal Processing}\label{bkgrnd:gsp}

In Graph Signal Processing, a node feature vector $\mathbf{x}\in\mathbb{R}^n$ is viewed as a signal on the vertices of a weighted, undirected graph $G=(V,E,w)$, $|V|=n$~\citep{shuman2013emerging,ortega2018graph}. Let $L=U\Lambda U^\top$ be its Laplacian with eigendecomposition $L\boldsymbol{\nu}_k=\lambda_k\boldsymbol{\nu}_k$, $0=\lambda_1\le\cdots\le\lambda_n$. The \emph{graph Fourier transform} is defined by projecting the singals onto these eigenvectors
$
\widehat{\mathbf{x}} = U^\top\mathbf{x}.$
The projection onto the first several eigenvectors (small $\lambda_k$) captures the smooth portion of the signal and the projection onto the later eigenvectors (large $\lambda_k$) capture oscillatory ones. Classical message‑passing GNNs act as low‑pass filters~\citep{bo2021beyond,nt2019revisiting}, effectively only keeping the projection onto the first several eigenvectors; DYMAG instead aggregates with waveforms spanning low, mid, and high bands. (Details in \Cref{appx:bkgrnd:gsp}.)  

\subsection{Heat and Wave Dynamics on a Graph}\label{sec:bckgrnd_heat_wave}

For $\alpha > 0$, we define the $\alpha$-fractional graph Laplacian by $L^\alpha := U \Lambda ^\alpha U^\mathsf{T}$.
For each $i\in V$, we let \(\delta_i\) denote the Dirac signal at $i$ given by: for $\forall k \in V$,
$
\delta_i(k)=1
$ if $i=k$, $\delta_i(k)=0$ otherwise.
We define
\begin{equation}
\label{eqn:heat_eq_definition}
    \smash{-L^\alpha u^{(i)}_H(v,t) = \partial_t u^{(i)}_H(v,t),\quad u^{(i)}_H(v,0)=\delta_i(v), \quad \text{(Heat)}}
\end{equation}
\begin{align}
\smash{-L^\alpha u^{(i)}_W(v,t) = \partial^2_t u^{(i)}_W(v,t),\quad u^{(i)}_W(v,0)=\delta_i(v), \quad \partial_t u^{(i)}_W(v,0)=c\delta_i(v),\quad \text{(Wave)}}\label{eqn:wave_eq_def}
\end{align}
as the heat and wave equations with a initial value $\delta_i$ (and initial veclocity $c\delta_i$ for wave). On a connected graph $G$, they admit closed‐form solutions:
\begin{small}
\begin{align}
\label{eqn:Heat}
u^{(i)}_H(v,t) &= \sum_{k=1}^n e^{-t\lambda_k^\alpha}\langle \boldsymbol{\nu}_k, \delta_i \rangle \boldsymbol{\nu}_k(v), \quad\text{and}\\
u^{(i)}_W(v,t) &=\sum_{k=1}^n \cos(\sqrt{\lambda_k^\alpha} t)\langle \boldsymbol{\nu}_k,\delta_i \rangle \boldsymbol{\nu}_k(v)  + t\langle \boldsymbol{\nu}_1,c\delta_i \rangle\boldsymbol{\nu}_1(v) 
+\sum_{k=2}^n\frac{1}{\sqrt{\lambda_k^\alpha}}\sin(\sqrt{\lambda_k^\alpha} t)\langle \boldsymbol{\nu}_k,c\delta_i\rangle\boldsymbol{\nu}_k(v). \label{eqn:Wave}
\end{align}
\end{small}
These expressions extend to disconnected graphs and, by Remark~1 of~\cite{chew2022geometric}, are invariant to the choice of Laplacian eigenbasis. (See \Cref{appendix:joyce_statement} for details).
\subsection{Chaotic Dynamics on a Graph}
Chaotic dynamics, describing systems that have aperiodic behavior and sensitivity to initial conditions~\citep{strogatz2018nonlinear}, can be modeled by the Sprott dynamics~\citep{sprott2008chaotic}:
\begin{small}
\begin{equation}
\smash[t]{\frac{d}{dt} u^{(i)}_S(v_k, t)= -b \cdot u^{(i)}_S(v_k, t)+ \tanh( \sum_{v_j \in \mathcal{N}(v_k)} c_{k,j} u^{(i)}(v_j, t)),\quad u_S(\cdot,0)=\delta_i,}
\label{eqn:sprott_dynamics}
\end{equation}
\end{small}
Solutions remain bounded for $b>0$. For \(b=0.25\), fully connected graphs with generic couplings or sparse graphs exhibit positive Lyapunov exponents (chaos). ~\citep{arnold2006lyapunov,sprott2008chaotic}. (Full details in \Cref{appx:bkgrnd:chaos}.)

\section{Methods}
\label{sec:Methods}
\label{sec:dymag}

DYMAG is graph neural network consisting of two main parts. 

\begin{enumerate}
\item \textbf{Waveform Bank Creation}: A diverse bank of multi-scale waveforms is constructed by solving the PDEs considered in Section \ref{sec:bg}. 
These waveforms define a set of basis functions that encode diverse  patterns across spatial and temporal scales. (See \Cref{sec:waveform_bank}.)
    
\item \textbf{Multi-scale Aggregations}: At each layer \(1\leq\ell\leq L\), node representations \(X^{(\ell-1)}\) are convolved with the waveform bank. 
The result is then passed through an MLP to produce an updated representation \(X^{(\ell)}\). This step replaces standard message passing mechanisms by aggregation via sophisticated, multiscale waveforms. (See \Cref{sec:msmp}.) 
\end{enumerate}


\begin{figure}[!htbp]
\centering
    \scalebox{1.0}{
    \begin{minipage}{0.43\textwidth}
    \centering
    {\scriptsize (a) \textsc{WaveformCreation} }
    \includegraphics[width=\linewidth]{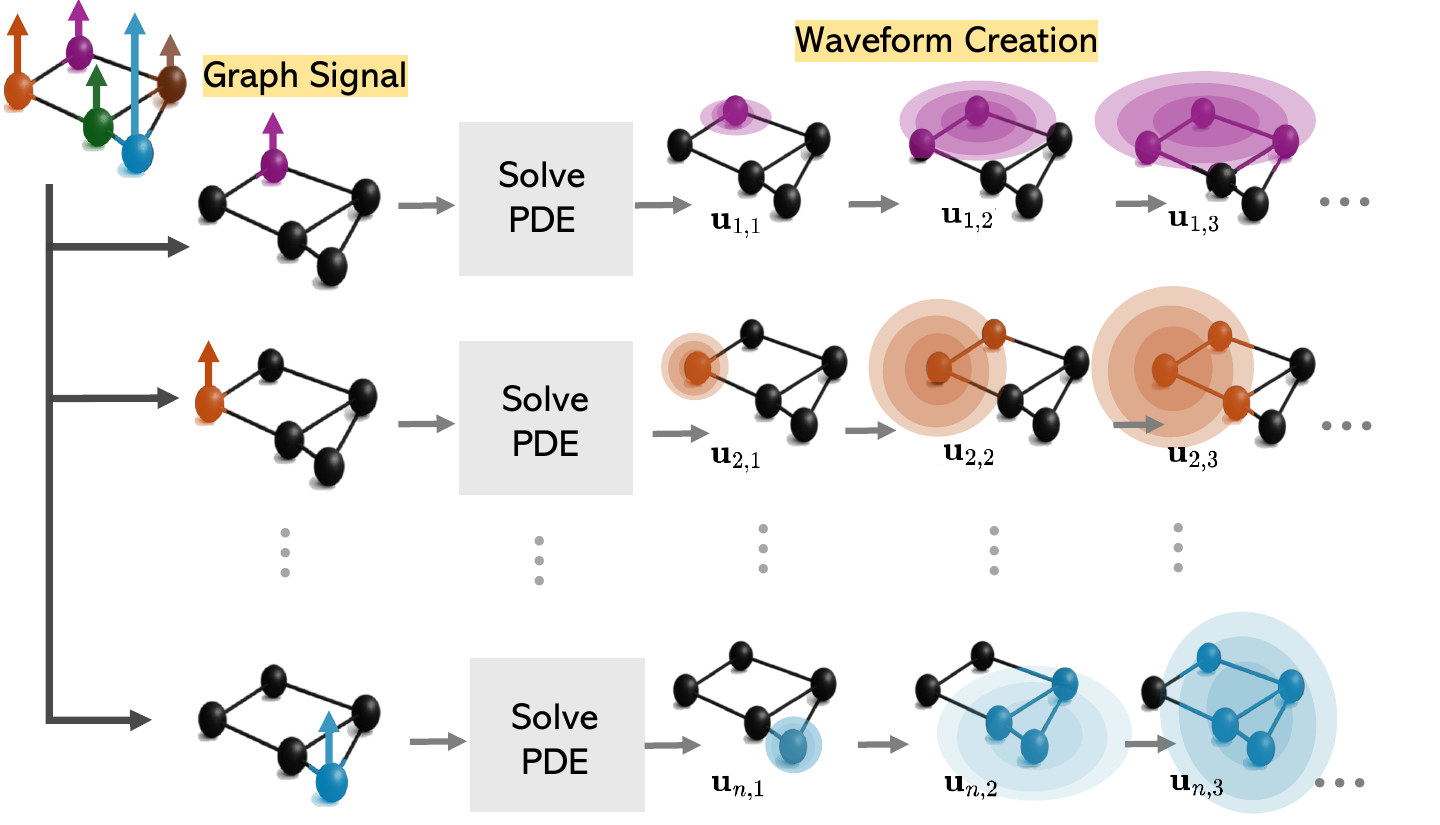}
    \end{minipage}
    \hfill
    \begin{minipage}{0.38\textwidth}
    \centering
    {\scriptsize (b) \textsc{MultiscaleAggregation} }
    \includegraphics[width=\linewidth]{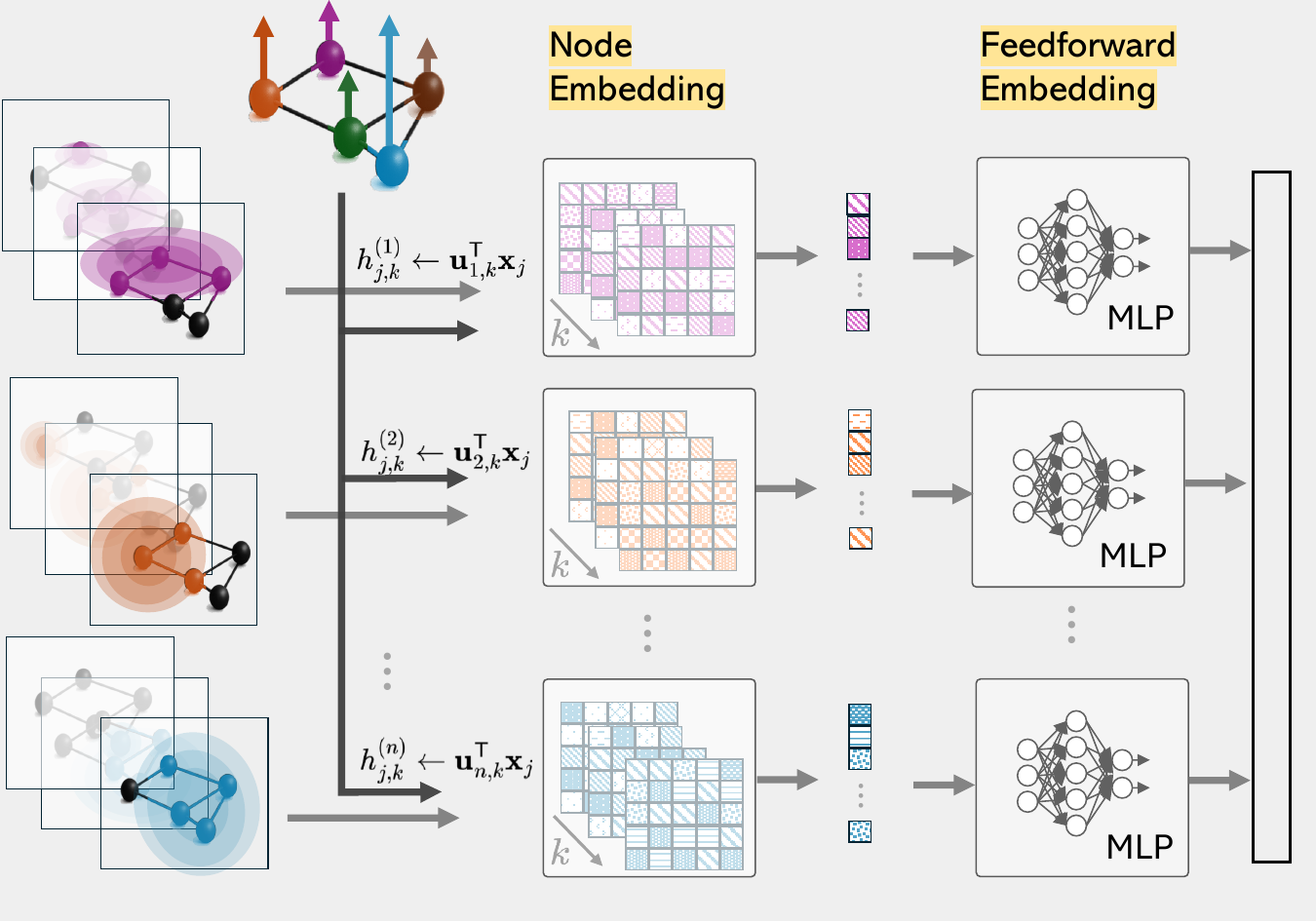}
    \end{minipage}
    \hfill
    \begin{minipage}{0.17\textwidth}
    \centering
   {\scriptsize (c) \textsc{DYMAG} }
    \includegraphics[width=\linewidth]{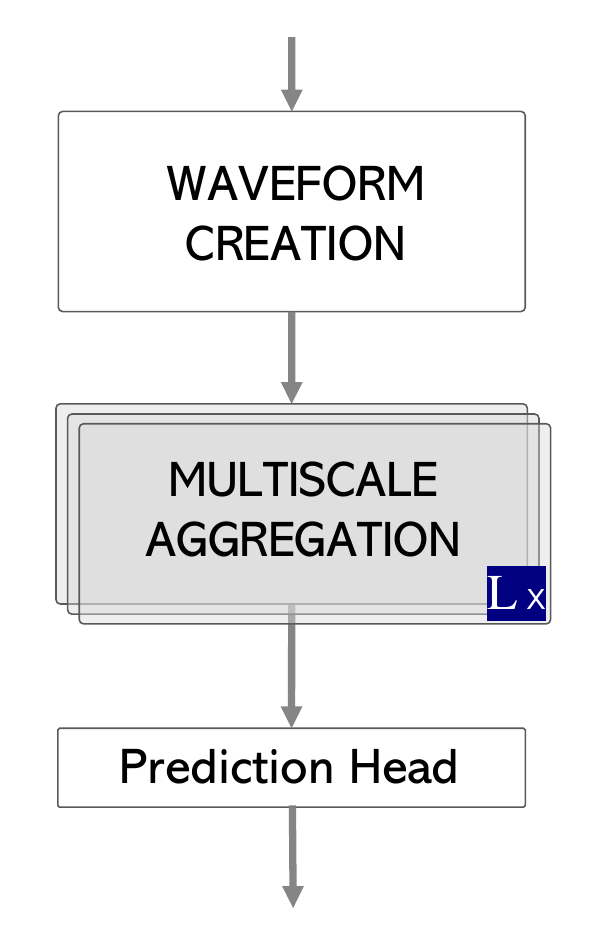}
    \end{minipage}
    }
\caption{\small \textbf{Visual illustration of DYMAG} (a) Waveform bank creation solving PDEs. (b) Multiscale Aggregation by taking inner product with the waveforms. (c) DYMAG consists of stacked layers and a prediction head.
}
\label{fig:dymag_combined}
\end{figure}






\begin{figure*}[t]
\centering

\begin{minipage}[t]{0.3\textwidth}
\begingroup
\SetAlCapFnt{\scriptsize}
\SetAlCapNameFnt{\scriptsize\bfseries}
\SetNlSty{}{\scriptsize}{}
\SetAlgoNlRelativeSize{-1}
\scriptsize

\begin{algorithm}[H]
\caption{\textsc{WaveformCreation}}
\label{alg:waveform_creation}
\KwIn{Graph $G=(V,E,w)$; sample times: 
$t_1,\dots,t_K$}
\KwOut{Waveforms $\mathcal U$}
\For{$v_i\in V,\;k=1,\dots,K$}{
  $\delta_i(\cdot)\gets\text{DiracSignals}(G,i)$\;
  $u^{(i)}(\cdot,t_k)\gets\text{SolvePDE}(\text{InitCond}=\delta_i)$\;
  $\mathbf u_{i,k}\gets u^{(i)}(\cdot,t_k)$\;
}
\Return{$\mathcal U=\{\mathbf u_{i,k}\}_{i\in V,\,1\le k\le K}$}
\end{algorithm}
\endgroup
\end{minipage}
\hfill
\hspace{0.01\textwidth}
\begin{minipage}[t]{0.3\textwidth}
\begingroup
\SetAlCapFnt{\scriptsize}
\SetAlCapNameFnt{\scriptsize\bfseries}
\SetNlSty{}{\scriptsize}{}
\SetAlgoNlRelativeSize{-1}
\scriptsize

\begin{algorithm}[H]
\caption{\textsc{MultiScaleAggr}}
\label{alg:mp}
\KwIn{Graph $G=(V,E,w)$; node features $X^{(\ell)}$; waveforms $\mathcal U$}
\KwOut{Updated features $X^{(\ell+1)}$}
\For{$v_i\in V,1\le j\le m,1\le k\le K$}{
  $h_{j,k}^{(i)}\gets \mathbf u_{i,k}^\top \mathbf x_j$\;
}
$\mathbf y_i\gets \mathrm{MLP}(\mathrm{vec}(h_{j,k}^{(i)}))$\;
\Return{$X^{(\ell+1)}=(\mathbf y_1^\top,\dots,\mathbf y_n^\top)^\top$}
\end{algorithm}
\endgroup
\end{minipage}
\hfill
\hspace{0.01\textwidth}
\begin{minipage}[t]{0.34\textwidth}
\begingroup
\SetAlCapFnt{\scriptsize}
\SetAlCapNameFnt{\scriptsize\bfseries}
\SetNlSty{}{\scriptsize}{}
\SetAlgoNlRelativeSize{-1}
\scriptsize

\begin{algorithm}[H]
\caption{\textsc{DYMAG}}
\label{alg:dymag}
\KwIn{Graph $G=(V,E,w)$; node features $X=\{\mathbf x_j\}$; sample times $\mathcal T$; layers $L$}
\KwOut{Output $Y$}
$\mathcal U\gets\textsc{WaveformCreation}(G,\mathcal T)$\;
$X^{(0)}\gets X$\;
\For{$\ell=1,\ldots,L$}{
${X^{(\ell)}\gets\textsc{MultiScaleAggr}(X^{(\ell-1)},\mathcal U)}$\;
\vspace{-1.5em}
}
$Y\gets\textsc{Readout}(X^{(L)})$\;
\Return{$Y$}
\end{algorithm}
\endgroup
\end{minipage}
\end{figure*}

\subsection{Waveform Bank Creation Using PDEs}\label{sec:waveform_bank}
Let \(u^{(i)}(v,t)\) denote the solution to the chosen PDE dynamics (wave, heat, or Sprott equations) with initial condition \(u^{(i)}(\cdot,0) = \delta_i(\cdot)\), where \(\delta_i(j) = 1\) if \(j = i\) and \(0\) otherwise (a Dirac signal centered at node \(i\)). When applicable (for second-order dynamics), we also set \(\partial_t u(\cdot, 0) = c\,\delta_i(\cdot)\), with a fixed hyperparameter \(c\geq 0\). 
We choose $K$ time points $\mathcal T=(t_1,\dots, t_K)$ by fixing a maximal time $T$ and then setting $t_k=kT/K$. We then define 
$
\mathcal U=\left\{\mathbf{u}_{i,k}\right\}_{i \in V,1\leq k\leq K},
$
where $\mathbf{u}_{i,k}$ is the vector $\mathbf{u}_{i,k}:=u^{(i)}(\cdot,t_k)\in\mathbb{R}^n$.
We refer to $\mathcal{U}$ as the \textit{PDE waveform bank} (see Algorithm~\ref{alg:waveform_creation} and \Cref{fig:dymag_combined}a). Each waveform $\mathbf{u}_{i,k}$ is centered at node $i$ and corresponds to a snapshot of the PDE dynamics at time scale $t_k$. The bank $\mathcal{U}$ collects such waveforms across all nodes and multiple time scales, similar to wavelets but more flexible thanks to the diverse dynamics. \Cref{fig:waveforms} shows basic waveforms and more complex patterns created via combinations (from the MLP discussed below).  

We note that $\mathcal{U}$ can be computed offline prior to training for increased computational efficiency. Additionally, note that the waveforms can be computed efficiently via either Chebshev approximation or a Runge-Kutta scheme. We further discuss on complexity and scalability in \Cref{appx:complexity}.

\subsection{Multi-scale Aggregation}\label{sec:msmp}

In each layer, $\ell$, we assume that we are given an $n\times m_\ell$ feature matrix $X^{(\ell)}$ (where $X^{(0)}$ consists of the initial node features). We let $\mathbf{x}_j\in\mathbb{R}^n$ denote the $r$-th column of $X^{(\ell)}$, which we interpret as a signal defined on $V$.
For each waveform $\mathbf u_{i,k}$ in the waveform bank $\mathcal{U}$
(Section~\ref{sec:waveform_bank}),
we perform an inner product with the node features, thought of as a convolution:
\begin{align}
  h_{j,k}^{(i)}
    &= \langle \mathbf u_{i,k},\mathbf x_j\rangle
       = \mathbf u_{i,k}^{\mathsf T}\mathbf x_j .
  \label{eq:inner-prod}
\end{align}
We then combine these convolved features by applying an MLP to the states $h^{(i)}_{j,k}$ associated with each node $v_i$, i.e., $\mathbf{y}_i=\text{MLP}\left(\text{vec}\left(h^{(i)}_{j,k}\right)\right)$. We then reorganize the $\mathbf{y}_i$ into a transformed feature matrix $X^{(\ell+1)} =(\mathbf{y}_1^{\mathsf{T}},\ldots,\mathbf{y}^{\mathsf{T}}_n)^{\mathsf{T}}$ (so that $\mathbf{y}^{\mathsf{T}}_i$ is the $i$-th row of $X^{(\ell+1)}$). See Algorithm~\ref{alg:mp} and \Cref{fig:dymag_combined}b.
 
The inner product, Eqn.~\ref{eq:inner-prod}, can also be interpreted as the feature $\mathbf{x}_j$ being updated via a message from a source node $v_i$ at scale $k$. Indeed,  message passing neural networks can be interpreted as performing such an inner product with a limited bandwidth rectangular waveform as shown in \Cref{fig:waveforms} and then applying the MLP. We remark that since we use a waveforms based on PDE solutions of various time snapshots, we obtain \emph{multi‑scale} embedding. As the time $t_k$ increases, the waveform effectively dilates and spreads to a larger neighborhood of vertices. Furthermore, via the MLP, DYMAG is able to learn novel combinations of the waveforms, either from different source nodes or at different time scales. This includes the diffusion wavelets \citep{coifman2006} which can be obtained by subtracting solutions to the heat equation at different time scales \citep{chew2022geometric}.


\paragraph{Downstream Readout}
After $L$ rounds of Multi-scale aggregation, the resulting node representations \( X^{(L)} = \{ \mathbf{x}_i^{(L)} \}_{i \in V} \) are used for prediction.  
For \emph{node-level tasks}, a shared MLP is applied independently to each node feature vector.  
For \emph{graph-level tasks}, node features are first aggregated using a permutation-invariant pooling operation (e.g., global mean or sum), followed by a task-specific MLP to produce the graph-level output. See  \Cref{fig:dymag_combined}c and Algorithm~\ref{alg:dymag}.

\subsection{Theoretical Properties Related to Dynamics-based Waveforms on the Graph}
\label{sec:Methods_motivation}
Below, we formulate properties of our waveforms and the information they are able to extract from the graph. 
These results serve as motivation for our method, which utilizes these dynamics as a novel aggregation paradigm for graph neural networks. 
Complete proofs are provided in \Cref{app: proofs}.
\subsubsection{Frequency Domain Characteristics of Waveform Based Message Aggregation}\label{sec: frequency characteristics}

Standard message passing can be viewed as convolving the node features with a single, simple, rectangular waveform.   
From the perspective of graph signal processing, this corresponds to a \emph{low-pass} filtering which only preserves the low-frequency (smooth) portion of the node features. 
In contrast, 
\textsc{DYMAG} employees a richer, more sophisticated bank of waveforms, which we will show allows DYMAG to extract a variety of different types of information. 

We first consider a  function defined by 
        $
    u_{\mathrm{BP}}^{(i)}(v,t)
        \;=\;
        u_H^{(i)}(v,t_1)\;-\;u_H^{(i)}(v,t_2)
        $
        for two fixed times, $0<t_1<t_2$. 
        In the graph-Fourier domain its response at frequency (eigenvalue) $\lambda_k$ is
        $
        e^{-t_1\lambda_k^\alpha}-e^{-t_2\lambda_k^\alpha}.
        $
        This function  
        (i) is $0$ at $\lambda_k=0$, 
        (ii) tends to $0$ as $\lambda_k\to\infty$, and 
        (iii) reaches a single maximum at
        $
\lambda^\star=\left(\frac{1}{t_2-t_1}\log\!\bigl(\tfrac{t_2}{t_1}\bigr)\right)^{1/\alpha}.
        $
Thus, $u_{BP}^{(i)}$ suppresses both very low and very high frequencies, but keeps information in a moderate frequency band (which depends on $t_1$ and $t_2$).       Therefore, we call $u^{(i)}_{BP}$ a band-pass function.
Notably, DYMAG has the ability to learn this function via the use of the MLP which is applied after Eqn.~\ref{eq:inner-prod}.
    We next consider the solution to the wave equation  given by Eqn.~\ref{eqn:Wave}, for simplicity focusing on the case  where $c=0$. The frequency response at each $\lambda_k$ is given by $\cos(\sqrt{\lambda_k^{\alpha}} t)$. Since this function peaks and falls in multiple different ``bands" we think of it as a \emph{multi-band-pass} function. This leads us to the following proposition. 


\begin{proposition}[Band-pass information]
\label{prop:bandpass}
DYMAG is able to extract band-pass, or even multi-band-pass information information from the node features.
\end{proposition}
\begin{proof}[Proof sketch] In heat-equation case, DYMAG can learn the band-pass function $u^{(i)}_{BP}$ via suitable weights in the MLP. In the wave equation case, DYMAG is able to capture multi-band-pass information as a consequence of the sinusoidal frequency responce of the wave solution, $u_W$.
\end{proof}


    In addition to the above propositions, we note that DYMAG is also able to learn low-pass information, similar to standard message passing networks. This is a direct consequence of the fact that $u_H^{(i)}$ has a decreasing frequency response $e^{-t\lambda_k^\alpha}$. It can also learn high-pass information via function $u^{(i)}_{\text{High}}=1-u^{(i)}_H$. We next discuss how the 
 frequency-domain characteristics of DYMAG help alleviate the following limitations of standard GNNs:
 
    \textbf{Over-smoothing}: Message passing networks 
    utilize rectangular pulse waveforms,  which act as low-pass filters, i.e., smoothing operators.  With each layer, the features get smoother and smoother, eventually become nearly constant, which  limits their usefulness. By contrast, DYMAG is able to learn band-pass, high-pass, and multi-band-pass information in addition to standard low-pass information. This allows it to avoid the oversmoothing problem. 
    
    \textbf{Under-reaching}: Message passing networks only aggregate within local, one-hop neighborhoods. Thus their receptive field is equal to the number of layers, which must be kept small to avoid severe oversmoothing. This limits their ability to capture global structure or long range interactions. DYMAG, on the other hand performs aggregation via waveforms which are not confined to one-hop neighborhoods and is able to capture global structure. 
        
    \textbf{Heterophily}: The local averaging operation in message passing networks, 
    are particularly problematic on heterophilic graphs where many nodes have different labels than their neighbors. DYMAG's diverse waveform banks are able to capture band-pass, multi-band-pass, and high-frequency information (in addition to low-pass).  This makes them well-suited to heterophilic graphs. Additionally, we note that  our experiment shows that the Sprott dynamics perform particularly well on node classificaiton on heterophilic graphs (see \Cref{fig:bar}), perhaps because of their ability to detect subtle changes in different portions of the network structure. 
\subsubsection{General Properties of Solutions}
The following result shows that DYMAG is able to identify the connected components of $G$. 


\begin{proposition}[Identification of Connected Components]
\label{prop:connected_components informal}
Let $u^{(i)}(v,t)$ denote the solution to the heat equation, 
wave equation, 
or Sprott chaotic dynamics. 
Suppose that $G$ is not connected.
Then, for any $v$ which is not in the same connected component as $v_i$, and all $t\geq0$, we have 
$u^{(i)}(v,t) = 0.$
\end{proposition}



\begin{proof}[Proof sketch]
We verify that $\tilde{u}^{(i)}(v,t) := u^{(i)}(v,t) \cdot \mathbbm{1}_{v \in \mathcal{C}}$, where $\mathcal{C}\subseteq V$ is the component containing $v_i$ 
satisfies the same PDE 
as $u^{(i)}(v,t)$. The result follows by uniqueness of solutions.
\end{proof}

Due to Proposition \ref{prop:connected_components informal}, we will assume that $G$ is connected in the following sections. However, we note that many of the results still apply to disconnected graphs with suitable modifications.

\subsubsection{Heat Dynamics}

The continuous and global nature of \modelname{} allows it to instantaneously have a receptive field over the entire graph. Intuitively, this corresponds to information spreading instantaneously over the  graph (although for small values of $t$ the energy $u^{(i)}_H(v,\cdot)$ will be mostly concentrated  near $v_i$). This is in contrast to message passing networks where the receptive field about each node is equal to the number of layers (which is usually must be kept small in order to avoid over-smoothing).

\begin{proposition}\label{prop: full support}
Let $G$ be connected and let $L$ be the random-walk Laplacian $L_{rw}$ (with $\alpha=1)$. Let $u^{(i)}_H(v,t)$ be the solution to the heat equation, Eqn.~ \ref{eqn:Heat}. 
Then $u^{(i)}_H(v,t)>0$ for all $v\in V$ and $t>0$.
\end{proposition}
\begin{proof}[Proof Sketch] This is a consequence of a relationship between $u^{(i)}_H$ and continuous-time random walks established in \Cref{thm: RW formal}.  
\end{proof}

Our next two results analyze the energy decay of $u_H^{(i)}$. They suggests that graphs with a larger $\lambda_2$ will have a faster rate of energy decay. The second eigenvalue of a graph can be related to the isoperimetric ratio of a graph through Cheeger's inequality, thereby revealing information on graph structure and how ``bottlenecked" a particular graph is \citep{spielman2019spectral}.
Additionally, they show that the properties of heat energy decay can distinguish between between graph structures. We note that although the assumptions for Proposition \ref{prop: heat between} represents a rather specific set of conditions, we expect that when two graphs have edges generated according to a similar rule or distribution, the more densely connected graph will have more rapidly decaying heat energy.

\begin{proposition}[Heat energy]\label{prop: heat energy short} Let $G$ be connected, and
let $u^{(i)}_H(v,t)$ be as in the solution to the heat equation with initial condition $\delta_i$ as in ~\Cref{eqn:Heat}.
Then, 
    $
        e^{-2t \lambda_n^\alpha} \leq \| u^{(i)}_H(\cdot, t)\|_2^2 \leq |\boldsymbol{\nu}_1(i)|^2 + e^{-2t\lambda_2^\alpha}. 
    $
\end{proposition}

\begin{proof}[Proof sketch] It follows from
    {\small $\|u^{(i)}_H(\cdot, t) \|_2^2 = \sum_{k = 1}^n e^{-2t\lambda_k^\alpha} |\langle \boldsymbol{\nu}_k, \delta_i \rangle|^2 = \sum_{k = 1}^n e^{-2t\lambda_k^\alpha} |\boldsymbol{\nu}_k(i)|^2. \qedhere$}
\end{proof}


\begin{proposition}[Heat energy between graphs]
\label{prop: heat between}
Let $G$ and $G'$ be graphs on $n$ vertices with fractional Laplacians $L_G^\alpha$ and $L_{G'}^\alpha$ and let $\delta_i$ and $\delta_{i'}$ be initial conditions for \Cref{eqn:heat_eq_definition} on $G$ and $G'$. Assume: 
%
(i) $L_{G'}^\alpha \succcurlyeq L_G^\alpha$, i.e.,  $\mathbf{v}^\mathsf{T} L_{G'}^\alpha \mathbf{v} \geq \mathbf{v}^\mathsf{T} L_{G}^\alpha \mathbf{v}\:\:\text{for all }\mathbf{v}\in\mathbb{R}^n,$
(ii) We have $| \boldsymbol{\nu}_k'(i)|^2 \leq (1 + \eta_k(t)) |\boldsymbol{\nu}_k(i)|^2$  for all $1\leq k \leq n,$ where we also assume $\eta_k(t) := \exp(2t ((\lambda_k')^\alpha-\lambda_k^\alpha)) - 1 \geq 0.$

Then, with $u_H$ and $u_H'$ defined as in \Cref{eqn:Heat},
we have 
$\|(u^{(i)}_H)'(\cdot,t)\|^2_2 \leq \|u^{(i)}_H(\cdot, t)\|_2^2.$
\end{proposition}
\begin{proof}[Proof sketch]
The result is a consequence of Parseval's identity.
\end{proof}
Finally, we restate some known results that provide additional foundation linking the behavior of the heat equation solutions to graph topology. Lemma 1 of \citet{Crane2017} shows that the heat equation encodes shortest path distances $d(v_i,v_j)$ between nodes on the graph: 


\begin{proposition}[Relation to distances, (Lemma 1 of \citet{Crane2017})]\label{prop:dist}
Let $u^{(i)}_H$ denote the solution to Eqn.~\ref{eqn:heat_eq_definition} with initial condition $\delta_i$ (and $\alpha=1)$. Then,   
    $d(v_i, v_j) = \lim_{t\to 0} \frac{\log u_H^{(i)}(v_j, t) }{\log t}.
$
\end{proposition} 
 We next consider, the Ollivier-Ricci curvature. This is a discrete notion of curvature, meant to parallel the tradition notion of Ricci curvature in Riemmanian geometry. It is defined by  
$\kappa(v_i, v_j) = 1 - W_1(\mu_{v_i}, \mu_{v_j})/d(v_i ,v_j),$ where $\mu_v$ is a probability measure centered around $v$ (see \cite{munch2019} for details), $W_1$ the 1-Wasserstein distance, and $d(v_i,v_j)$ is the distance (shortest path length) from $v_i$ to $v_j$. 
The following result from \citet{munch2019} relates $\kappa$ to the  heat equation.

\begin{proposition}[Relation to Ollivier-Ricci curvature, (Theorem 5.8 of \cite{munch2019})]
\label{prop:ollivier_ricci}
Let $L=D-A$ be the unnormalized Laplacian, then 
    {\small $\kappa(v_i, v_j) = \lim_{t\to 0^+} \frac{1}{t} \bigg(1 - \frac{W_1(u^{(i)}_H(\cdot, t), u^{(j)}_H(\cdot, t))}{d(v_i,v_j)} \bigg).$}
\end{proposition}

\subsubsection{Wave Dynamics}

The periodicity of the solution to the wave equation endows it with the ability to capture long range interactions. Central to this argument is the wave energy, analyzed in the following proposition which focuses on the case where the initial velocity $c$ is equal to zero:

\begin{proposition}[Wave energy bounds]
\label{prop: wave energy}
Let $u_W^{(i)}(v,t)$ be the solution to the fractional wave equation Eqn.~\ref{eqn:Wave} with initial conditions $u_W^{(i)}(\cdot,0) = \delta_i$ and $\partial_t u_W^{(i)}(\cdot,0) = 0$. Then, for any time $t \geq 0$, the energy of the waveform satisfies
$|\boldsymbol{\nu}_1(i)|^2 \leq \| u_W^{(i)}(\cdot,t) \|_2^2 \leq  1.$
\end{proposition}

\begin{proof}[Proof sketch]
By Parseval's identity and the explicit solution in \Cref{eqn:Wave}, we expand:
$\| u_W^{(i)}(\cdot,t) \|_2^2 = \sum_{k=1}^n \cos^2\left(\sqrt{\lambda_k^\alpha}t\right) |\langle \boldsymbol{\nu}_k, \delta_i \rangle|^2=\sum_{k=1}^n \cos^2\left(\sqrt{\lambda_k^\alpha}t\right) |\boldsymbol{\nu}_k(i)|^2.$
Since $\cos^2(\cdot) \in [0,1]$ and $\sum_k |\langle \boldsymbol{\nu}_k, \delta_i \rangle|^2 = \| \delta_i \|_2^2 = 1$, the result follows.
\end{proof}

This shows that, unlike heat kernels (which decay over time), the wave energy oscillates, retaining signal over time, and thus can reflect non-local interactions such as those created by cycles in the graph. These oscillations allow the waveforms to “echo” through the graph and revisit distant parts of the structure - a behavior well-suited for recovery of topological features.

\begin{proposition}[Recovery of eigenspectrum from waveforms]
\label{prop:wave_eigenvalue}
Let $G$ be connected, and let $u_W^{(i)}(v,t)$ be the solution to the fractional wave equation (\Cref{eqn:Wave}), with initial conditions $u_W^{(i)}(\cdot,0) = \delta_i$ and $\partial_t u_W^{(i)}(\cdot,0) = 0$. Then, for any fixed node $v$, the sequence of values $u_W^{(i)}(v, t_1), \dots, u_W^{(i)}(v, t_m)$ obtained from time samples can be used to approximate the full Laplacian eigenspectrum $\{ \lambda_k^\alpha \}_{k=1}^n$ up to arbitrary precision, provided sufficient time resolution.
\end{proposition}

\begin{proof}[Proof sketch]
$u_W^{(i)}(v, t)$ is a linear combination of cosines  at frequencies $\sqrt{\lambda_k^\alpha}$$u_W^{(i)}(v, t) = \sum_{k=1}^n \cos\left(\sqrt{\lambda_k^\alpha}t\right) \langle \boldsymbol{\nu}_k, \delta_i \rangle \boldsymbol{\nu}_k(v)$.
Thus, we may apply  the Shannon-Nyquist sampling  theorem.
\end{proof}

The graph eigenspectrum encodes a wide range of graph invariants and properties. Proposition \ref{prop:wave_eigenvalue} demonstrates that the solutions to the wave equation relate to graph spectral properties, and that this entire information is contained in the solutions of the wave equation \textit{at each node}. For example, we have the following  corollary:

\begin{corollary}[Cycle Length]
\label{cor:cycle_length}
    The size of cycle graph $C_n$ can be determined from the solution to the fractional wave equation at a single node $v$.
\end{corollary}
\begin{proof}[Proof sketch]
The result follows from Proposition \ref{prop:wave_eigenvalue} and the fact that the length of a cycle graph is contained in its eigenspectrum.
\end{proof}

More generally, when the graph is not a cycle but contains a cyclic subgraphs as a prominent topological feature, this proposition provides some intuition for why the wave-equation is well suited to pick up that a node belongs to a cycle and recover dimension 1 homology.


\section{Empirical Results}
\label{sec:empirical}


We evaluate \modelname{} across diverse tasks to assess its ability to recover geometric/topological structure and generalize to downstream biological, chemical, and materials applications. In this section, we set $\alpha=1$ so that the fractional Laplacian coincides with the ordinary graph Laplacian. We conduct some experiments with other exponents $\alpha$ in \Cref{appendix:fractional}. Baselines include message-passing GNNs (MPNN~\citep{gilmer2017neural}, 
GAT~\citep{velickovic2017graph}, GIN~\citep{xu2018powerful}), diffusion-based methods (GRAND~\citep{chamberlain2021grand}, GRAND\texttt{++}\citep{thorpe2022grand++}), and  GraphGPS~\citep{rampavsek2022recipe}, a state-of-the art graph transformer. We also compare against a GNN built with fixed-scale wavelets (GWT~\citep{coifman2006}) and a neural approximation algorithm of the extended persistence diagram (EPD)
~\citep{yan_neural_2022}. For molecular and materials prediction, we include pretrained models such as ProtBERT~\citep{brandes2022proteinbert}, MolBERT~\citep{fabian2020molecular}, and GeoCGNN~\citep{cheng2021geometric}.

In all tables, \best{best} and \second{second best} results are highlighted. Ties within one standard deviation are treated as equivalent. Unless noted otherwise, results reflect means over 10-fold cross-validation. Implementation details, including hyperparameter selection, Chebyshev approximations, and complexity analysis, are provided in \Cref{sec:appendix_implementation}. Full results with standard deviations appear in \Cref{app:more_experiments}.

\subsection{Geometric and Topological Properties}
\label{sec: results_geom_and_top}
\begin{wrapfigure}{r}{0.5\textwidth}
  \centering
  \vspace{-2ex}  
  \includegraphics[width=0.5\textwidth]{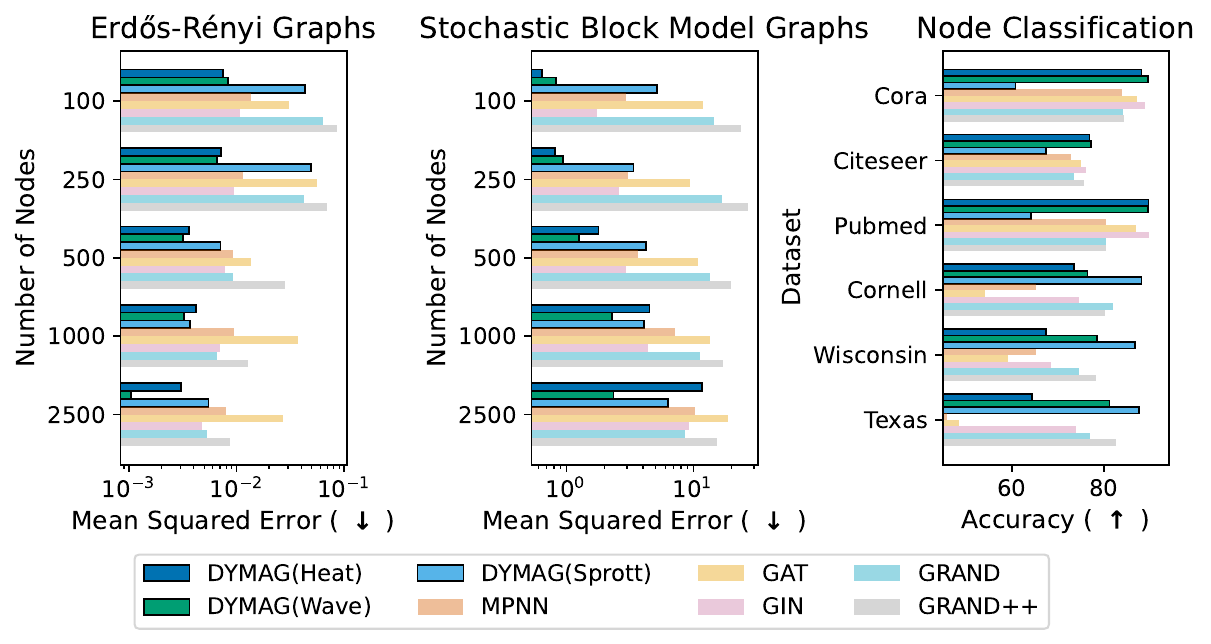}
  \caption{\small Mean squared error (MSE, lower is better) for predicting the generating parameters of random graphs and node classification accuracy (higher is better) for homophilic and heterophilic datasets. (See also \Cref{tab:results-graph-params,tab:results-node-classification} in the Appendix.)}
  \label{fig:bar}
  \vspace{-1ex} 
\end{wrapfigure}
We first evaluated the expressivity of multiscale dynamics as a replacement for message passing by training \modelname{} to recover geometric and topological features, including Ollivier–Ricci curvature and the persistence image representation~\citep{adams_persistence_2016} of the extended persistence diagram. Ollivier–Ricci curvature~\citep{ollivier_ricci_2007}, a discrete analog of Ricci curvature from Riemannian geometry, captures local graph geometry. For persistent homology, diagrams are computed from ascending and descending filtrations of each node’s 5-hop neighborhood, using node degree as the filter~\citep{cohen-steiner_extending_2009,yan_neural_2022}. The resulting persistence images encode connected components (0 homology) and loops (1st homology).

We evaluated the model on Erd\H{o}s-R\'enyi graphs \citep{erdos1960evolution}, $G(n,p)$, with $n\in \{100,200\}$  and $p\in\{0.04,0.06,0.08\}$, stochastic block model (SBM) graphs, and several citation graphs.
The results are shown in \Cref{tab:merged_results_noER} and \Cref{tab:merged_results} (appendix). \modelname{} significantly improves prediction accuracy on both Ollivier–Ricci curvature and persistent homology compared to standard GNNs and GRAND. For persistent homology, \modelname{} performs on par with the purpose-built model of~\citet{yan_neural_2022}, a neural approximation of the Union-Find EPD  algorithm, despite DYMAG not being designed specifically for this task. \modelname{} generalizes well across both synthetic and real-world graphs, including Cora~\citep{mccallum_automating_2000}, Citeseer~\citep{giles_citeseer_1998}, and PubMed~\citep{sen_collective_2008}. Full Results on ER graphs and additional experiments using alternative filtrations are provided in \Cref{app:more_experiments}.
We note that while persistent homology can be computed directly, it is computationally expensive with cost $\mathcal{O}(g^3)$, $g$ being the number of generators~\citep{zomorodian2004computing}. More importantly, these experiments highlight that \modelname{} learns rich graph representations that capture topological features when useful for prediction, and it can adapt to the task to extract relevant information for regression or classification.

As shown in \Cref{fig:bar} and \Cref{tab:results-graph-params,tab:results-node-classification} (appendix), we evaluate expressivity via two tasks as proxies: recovering parameters of random graphs and node classification on homophilic (Pubmed, Citeseer, Cora) and heterophilic (Texas, Wisconsin, Cornell)~\citep{pei2020geom} networks. We see that the heat and wave versions of DYMAG are the top performing models on most of the data sets whereas the Sprott version of DYMAG is the top performing model on the heterophilic data sets, possibly because these data sets need a model which is sensitive to small changes in local graph structure. 
\Cref{app:more_experiments} presents results on node-level Ollivier–Ricci curvature, persistent homology with alternative filtrations, and an analysis of fractional heat and wave dynamics across different $\alpha$ values.%
\vspace*{-1em}
\begin{table*}[!htbp]
\scriptsize
\caption{\small
Mean squared error (MSE) for predicting Ollivier–Ricci curvature ($\kappa$) and extended persistence images (EP) across citation and OGBN graph benchmarks.  Results are averaged over 10 runs; lower is better.  ``–'' indicates not applicable, timeout or out of memory. The Results on ER graphs are provided in \Cref{app:more_experiments}.}
\label{tab:merged_results_noER}
\renewcommand{\arraystretch}{1.1}
\centering
\resizebox{\textwidth}{!}{%
\begin{tabular}{l
  |cc|cc|cc|c|c
}
\toprule
\multirow{2}{*}{\textbf{Model}}
  & \multicolumn{2}{c|}{\textbf{Cora}}
  & \multicolumn{2}{c|}{\textbf{Citeseer}}
  & \multicolumn{2}{c|}{\textbf{PubMed}}
  & \multicolumn{1}{c|}{\textbf{ogbn-mag}}
  & \multicolumn{1}{c}{\textbf{ogbn-papers100M}}
\\
  & $\kappa$ & EP
  & $\kappa$ & EP
  & $\kappa$ & EP
  & EP   & EP
\\
\midrule
\modelname\textsubscript{(Heat)}
  & \best{1.73e-2} & \second{1.45e-4}
  & \best{2.09e-2} & \second{3.94e-4}
  & \second{7.30e-3} & \second{7.93e-3}
  & \second{5.39e-2} & \second{9.03e-2} \\

\modelname\textsubscript{(Wave)}
  & \second{1.85e-2} & \best{7.41e-5}
  & \second{3.41e-2} & \best{1.58e-4}
  & \best{6.51e-3}   & \best{3.29e-3}
  & \best{8.01e-3}   & \best{3.25e-2} \\

\modelname\textsubscript{(Sprott)}
  & 1.34e-1 & 3.51e-3
  & 1.46e-1 & 7.64e-3
  & 2.97e-2 & 8.96e-2
  & –       & –         \\
\midrule
MPNN
  & 2.40e-1 & 2.72e-3
  & 2.20e-1 & 6.83e-3
  & 1.69    & 4.29e-1
  & 2.93e-1 & 6.19e-1 \\

GAT
  & 7.36e-1 & 5.04e-2
  & 9.04e-1 & 2.93e-2
  & 1.55e-1 & 4.79e-1
  & 4.47e-1 & 8.03e-1 \\

GIN
  & 1.56e-1 & 5.53e-3
  & 1.72e-1 & 3.94e-3
  & 8.34e-3 & 1.27e-1
  & 2.49e-1 & 6.43e-1 \\

GWT
  & 4.07e-2 & 6.79e-4
  & 3.58e-2 & 9.12e-4
  & 9.15e-3 & 2.16e-2
  & 5.83e-2 & 2.71e-1 \\

GraphGPS
  & 4.41e-1 & 2.71e-2
  & 2.82e-1 & 3.16e-2
  & 7.68e-1 & 1.94e-1
  & 3.78e-1 & 4.05e-1 \\

GRAND
  & 6.81e-2 & 8.13e-4
  & 1.24e-1 & 6.87e-4
  & 2.73e-2 & 3.37e-2
  & –       & –         \\

GRAND\texttt{++}
  & 1.74e-1 & 8.16e-3
  & 3.09e-1 & 4.21e-3
  & 8.72e-2 & 3.07e-1
  & –       & –         \\
\midrule
Neural EPD Approx.
  & –       & 5.80e-5
  & –       & 1.29e-4
  & –       & 2.74e-3
  & 6.73e-3 & 3.18e-2 \\
\bottomrule
\end{tabular}%
}
\end{table*}
\begin{table*}[!htbp]
\caption{
\small Performance of \modelname{} on four datasets: PROTEINS, DrugBank, Materials Project (MP), and the DTS AIDS Antiviral Screen. We report R$^2$ for the first three and balanced accuracy for the Antiviral Screen. Results are mean $\pm$ std over 10‑fold CV.
}
\scriptsize
\label{tab:results-mols}
\centering
\resizebox{\textwidth}{!}{%
\begin{tabular*}{1.0\textwidth}{@{\extracolsep{\fill}}c|c|cc|c|c}
\toprule
\multirow{2}{*}{\textbf{Model}}
  & \multicolumn{1}{c|}{\textbf{PROTEINS}}
  & \multicolumn{2}{c|}{\textbf{DrugBank}}
  & \textbf{MP}
  & \textbf{Antiviral Screen} \\
& Dihedral Angles & TPSA & \# Aromatic Rings & Band Gap & Active/Inactive \\
\midrule
\modelname\textsubscript{ (Heat)}
  & $\bestmm{0.89 \pm 0.01}$
  & $\bestmm{0.97 \pm 0.01}$
  & $\bestmm{0.97 \pm 0.02}$
  & $\bestmm{0.61 \pm 0.03}$
  & $0.54 \pm 0.02$ \\
\modelname\textsubscript{ (Wave)}
  & $\secondmm{0.81 \pm 0.03}$
  & $\secondmm{0.90 \pm 0.01}$
  & $\secondmm{0.88 \pm 0.01}$
  & $\secondmm{0.55 \pm 0.02}$
  & $\secondmm{0.61 \pm 0.01}$ \\
\modelname\textsubscript{ (Sprott)}
  & $0.76 \pm 0.01$
  & $0.77 \pm 0.01$
  & $0.82 \pm 0.03$
  & $\secondmm{0.54 \pm 0.03}$
  & $\bestmm{0.63 \pm 0.02}$ \\
\hline
MPNN
  & $0.78 \pm 0.01$
  & $0.71 \pm 0.01$
  & $0.81 \pm 0.01$
  & $0.37 \pm 0.05$
  & $0.51 \pm 0.02$ \\
GAT
  & $0.72 \pm 0.02$
  & $0.78 \pm 0.02$
  & $0.83 \pm 0.02$
  & $0.40 \pm 0.03$
  & $\secondmm{0.59 \pm 0.03}$ \\
GIN
  & $0.69 \pm 0.03$
  & $0.69 \pm 0.01$
  & $0.77 \pm 0.01$
  & $0.38 \pm 0.03$
  & $\secondmm{0.60 \pm 0.02}$ \\
GWT
  & $\secondmm{0.81 \pm 0.02}$
  & $0.83 \pm 0.02$
  & $0.85 \pm 0.01$
  & $0.42 \pm 0.02$
  & $0.58 \pm 0.02$ \\
GraphGPS
  & $0.64 \pm 0.03$
  & $0.63 \pm 0.02$
  & $0.67 \pm 0.04$
  & $0.31 \pm 0.02$
  & $0.54 \pm 0.03$ \\
GRAND
  & $0.76 \pm 0.03$
  & $0.53 \pm 0.04$
  & $0.64 \pm 0.03$
  & $0.27 \pm 0.03$
  & $0.49 \pm 0.03$ \\
GRAND\texttt{++}
  & $0.62 \pm 0.03$
  & $0.56 \pm 0.02$
  & $0.61 \pm 0.02$
  & $0.31 \pm 0.02$
  & $0.51 \pm 0.02$ \\
\hline
Pretrained
  & $\secondmm{0.83 \pm 0.03}$
  & $\bestmm{0.98 \pm 0.01}$ & $\bestmm{0.97 \pm 0.01}$ & $\bestmm{0.62 \pm 0.05}$ & $\secondmm{0.59 \pm 0.02}$ \\
  Model
  & (ProtBERT) 
  & (MolBERT) & (MolBERT) & (GeoCGNN) & (ProtBERT) \\
\bottomrule
\end{tabular*}%
}
\end{table*}
\subsection{Proteins, Molecules, and Materials}
\label{sec: results_molecules}
We evaluated \modelname{} on graphs representing proteins, drug-like molecules, and materials shown in \Cref{tab:results-mols,tab:classification} (appendix). We note that DYMAG's strong performance on these data sets indicates its potential positively impact society by furthering the design of materials, drugs, or other healthcare treatments.
More specifically, the tasks include predicting geometric and chemical properties such as dihedral angles, total polar surface area (TPSA), the number of aromatic rings, band gaps in materials, and anti-HIV activity. Datasets include PROTEINS~\citep{dobson_distinguishing_2003}, DrugBank~\citep{wishart_drugbank_2018}, the Materials Project~\citep{10.1063/1.4812323}, and the AIDS Antiviral Screen~\citep{nih_aids_screen}. Comparisons span standard message-passing GNNs (MPNN, GAT, GIN), diffusion-based models (GRAND, GRAND\texttt{++}), spectral methods (GWT), transformer-based models (GraphGPS), and domain-specific pre-trained models: ProtBERT~\citep{brandes2022proteinbert} for proteins, MolBERT~\citep{fabian2020molecular} for small molecules, and GeoCGNN~\citep{cheng2021geometric} for materials.
Across all datasets, \modelname{} consistently outperforms standard GNNs, GRAND, GRAND\texttt{++}, GraphGPS, and GWT by a wide margin. It matches the performance of powerful, task-specific pretrained models within 1 standard deviation overall, and significantly surpasses them on the PROTEINS and AIDS datasets.

Overall, the heat and wave versions of \modelname{} perform strongly across all molecular and material prediction tasks.
The Sprott (chaotic) variant shows more variable performance, which may reflect its heightened sensitivity to local graph structure. This behavior appears beneficial in settings with recurring structural motifs, such as the Antiviral Screen dataset, and may be less advantageous in tasks where such sensitivity is less critical. Additional results, including accuracy and training time, are provided in \Cref{app:more_experiments}.
\vspace{-0.1em}
\section{Conclusion}\label{sec:discussion}
\vspace{-0.1em}
We introduce DYMAG as a method for improving aggregation in GNNs. We use dynamics to generate a diverse bank of waveforms that span multiple frequency bands. Messages are aggregated by taking each node feature, interpreted as a graph signal, and projecting it onto this bank via inner products, producing a set of features that encode multi-scale information. The expressiveness of this representation, arising from the rich frequency structure of the waveforms, helps mitigate common GNN limitations such as oversmoothing, underreaching, and heterophily.

One limitation of our method is that the Sprott model is extremely sensitive to the graph structure. This is useful for some tasks which require detecting minute changes in structure. However, this be undesirable in other settings where one may want similar representations of nearly isomorphic.
Another  limitation is that our method is currently only applicable to supervised tasks. Extending DYMAG to unsupervised tasks, such as clustering, denoising,  or signal reconstruction could could be  in interesting avenue of future work.
Overall, DYMAG establishes a theoretically grounded, empirically strong message aggregation paradigm; future work will broaden its application to diverse graph tasks and refine the accompanying speed-up techniques.

\section*{Acknowledgments}
This research was partially funded and supported by CIFAR AI Chair [G.W.], NSERC Discovery grant 03267 [G.W.], NIH grants (1F30AI157270-01, R01HD100035, R01GM130847, R01GM135929) [G.W.,S.K.], NSF Career grant 2047856 [S.K.], NSF grant 2327211 [S.K., G.W., M.P.], NSF grant 2242769 [M.P.], NSF/NIH grant 1R01GM135929-01 [S.K.], NSF CISE grant 2403317 [S.K.], the Chan-Zuckerberg Initiative grants CZF2019182702 and CZF2019-002440 [S.K.], the Sloan Fellowship FG-2021-15883 [S.K.], the Novo Nordisk grant GR112933 [S.K.], Yale - Boehringer Ingelheim Biomedical Data Science Fellowship [D.B.], and Kavli Institute for Neuroscience Postdoctoral Fellowship [D.B.].





\bibliography{references}

\begin{thebibliography}{72}
\providecommand{\natexlab}[1]{#1}
\providecommand{\url}[1]{\texttt{#1}}
\expandafter\ifx\csname urlstyle\endcsname\relax
  \providecommand{\doi}[1]{doi: #1}\else
  \providecommand{\doi}{doi: \begingroup \urlstyle{rm}\Url}\fi

\bibitem[Rusch et~al.(2023)Rusch, Bronstein, and Mishra]{rusch2023survey}
T~Konstantin Rusch, Michael~M Bronstein, and Siddhartha Mishra.
\newblock A survey on oversmoothing in graph neural networks.
\newblock \emph{arXiv preprint arXiv:2303.10993}, 2023.

\bibitem[Keriven(2022)]{keriven2022not}
Nicolas Keriven.
\newblock Not too little, not too much: a theoretical analysis of graph (over) smoothing.
\newblock \emph{Advances in Neural Information Processing Systems}, 35:\penalty0 2268--2281, 2022.

\bibitem[Errica et~al.(2023)Errica, Christiansen, Zaverkin, Maruyama, Niepert, and Alesiani]{errica2023adaptive}
Federico Errica, Henrik Christiansen, Viktor Zaverkin, Takashi Maruyama, Mathias Niepert, and Francesco Alesiani.
\newblock Adaptive message passing: A general framework to mitigate oversmoothing, oversquashing, and underreaching.
\newblock \emph{arXiv preprint arXiv:2312.16560}, 2023.

\bibitem[Abu-El-Haija et~al.(2019)Abu-El-Haija, Perozzi, Kapoor, Alipourfard, Lerman, Harutyunyan, Ver~Steeg, and Galstyan]{abu2019mixhop}
Sami Abu-El-Haija, Bryan Perozzi, Amol Kapoor, Nazanin Alipourfard, Kristina Lerman, Hrayr Harutyunyan, Greg Ver~Steeg, and Aram Galstyan.
\newblock Mixhop: Higher-order graph convolutional architectures via sparsified neighborhood mixing.
\newblock In \emph{international conference on machine learning}, pages 21--29. PMLR, 2019.

\bibitem[Gao et~al.(2019)Gao, Wolf, and Hirn]{gao2019geometric}
Feng Gao, Guy Wolf, and Matthew Hirn.
\newblock Geometric scattering for graph data analysis.
\newblock In \emph{International Conference on Machine Learning}, pages 2122--2131. PMLR, 2019.

\bibitem[Perlmutter et~al.(2023)Perlmutter, Tong, Gao, Wolf, and Hirn]{perlmutter2023understanding}
Michael Perlmutter, Alexander Tong, Feng Gao, Guy Wolf, and Matthew Hirn.
\newblock Understanding graph neural networks with generalized geometric scattering transforms.
\newblock \emph{SIAM Journal on Mathematics of Data Science}, 5\penalty0 (4):\penalty0 873--898, 2023.

\bibitem[Gama et~al.(2019{\natexlab{a}})Gama, Bruna, and Ribeiro]{gama2018diffusion}
Fernando Gama, Joan Bruna, and Alejandro Ribeiro.
\newblock Diffusion scattering transforms on graphs.
\newblock January 2019{\natexlab{a}}.
\newblock 7th International Conference on Learning Representations, ICLR 2019 ; Conference date: 06-05-2019 Through 09-05-2019.

\bibitem[Gama et~al.(2019{\natexlab{b}})Gama, Ribeiro, and Bruna]{gama2019stability}
Fernando Gama, Alejandro Ribeiro, and Joan Bruna.
\newblock Stability of graph scattering transforms.
\newblock \emph{Advances in Neural Information Processing Systems}, 32, 2019{\natexlab{b}}.

\bibitem[Jiang et~al.(2024)Jiang, Zou, Zhang, and Wang]{jiang2024limiting}
Yuanhong Jiang, Dongmian Zou, Xiaoqun Zhang, and Yu~Guang Wang.
\newblock Limiting over-smoothing and over-squashing of graph message passing by deep scattering transforms.
\newblock \emph{arXiv preprint arXiv:2407.06988}, 2024.

\bibitem[Wenkel et~al.(2022)Wenkel, Min, Hirn, Perlmutter, and Wolf]{wenkel2022overcoming}
Frederik Wenkel, Yimeng Min, Matthew Hirn, Michael Perlmutter, and Guy Wolf.
\newblock Overcoming oversmoothness in graph convolutional networks via hybrid scattering networks.
\newblock \emph{arXiv preprint arXiv:2201.08932}, 2022.

\bibitem[Bodmann and Emilsd{\'o}ttir(2024)]{bodmann2024scattering}
Bernhard~G Bodmann and {\'I}ris Emilsd{\'o}ttir.
\newblock A scattering transform for graphs based on heat semigroups, with an application for the detection of anomalies in positive time series with underlying periodicities.
\newblock \emph{Sampling Theory, Signal Processing, and Data Analysis}, 22\penalty0 (2):\penalty0 17, 2024.

\bibitem[Xu et~al.(2024)Xu, Goldman, Guo, Hollander-Bodie, Trank-Greene, Adelstein, De~Brouwer, Ying, Krishnaswamy, and Perlmutter]{xu2023blis}
Charles Xu, Laney Goldman, Valentina Guo, Benjamin Hollander-Bodie, Maedee Trank-Greene, Ian Adelstein, Edward De~Brouwer, Rex Ying, Smita Krishnaswamy, and Michael Perlmutter.
\newblock {BLIS}-{N}et: Classifying and analyzing signals on graphs.
\newblock In \emph{Proceedings of The 27th International Conference on Artificial Intelligence and Statistics}, volume 238 of \emph{Proceedings of Machine Learning Research}, pages 4537--4545. PMLR, 02--04 May 2024.
\newblock URL \url{https://proceedings.mlr.press/v238/xu24c.html}.

\bibitem[Tong et~al.(2022)Tong, Wenkel, Bhaskar, Macdonald, Grady, Perlmutter, Krishnaswamy, and Wolf]{tong2022learnable}
Alexander Tong, Frederik Wenkel, Dhananjay Bhaskar, Kincaid Macdonald, Jackson Grady, Michael Perlmutter, Smita Krishnaswamy, and Guy Wolf.
\newblock Learnable filters for geometric scattering modules.
\newblock \emph{arXiv preprint arXiv:2208.07458}, 2022.

\bibitem[Hammond et~al.(2011)Hammond, Vandergheynst, and Gribonval]{hammond2011wavelets}
David~K Hammond, Pierre Vandergheynst, and R{\'e}mi Gribonval.
\newblock Wavelets on graphs via spectral graph theory.
\newblock \emph{Applied and Computational Harmonic Analysis}, 30\penalty0 (2):\penalty0 129--150, 2011.

\bibitem[Coifman and Maggioni(2006)]{coifman2006}
Ronald~R. Coifman and Mauro Maggioni.
\newblock Diffusion wavelets.
\newblock \emph{Applied and Computational Harmonic Analysis}, 21\penalty0 (1):\penalty0 53--94, 2006.
\newblock ISSN 1063-5203.
\newblock \doi{https://doi.org/10.1016/j.acha.2006.04.004}.
\newblock URL \url{https://www.sciencedirect.com/science/article/pii/S106352030600056X}.
\newblock Special Issue: Diffusion Maps and Wavelets.

\bibitem[McCallum et~al.(2000)McCallum, Nigam, Rennie, and Seymore]{mccallum_automating_2000}
Andrew~Kachites McCallum, Kamal Nigam, Jason Rennie, and Kristie Seymore.
\newblock Automating the {Construction} of {Internet} {Portals} with {Machine} {Learning}.
\newblock \emph{Information Retrieval}, 3\penalty0 (2), July 2000.
\newblock ISSN 1573-7659.
\newblock \doi{10.1023/A:1009953814988}.
\newblock URL \url{https://doi.org/10.1023/A:1009953814988}.

\bibitem[Giles et~al.(1998)Giles, Bollacker, and Lawrence]{giles_citeseer_1998}
C.~Lee Giles, Kurt~D. Bollacker, and Steve Lawrence.
\newblock {CiteSeer}: an automatic citation indexing system.
\newblock In \emph{Proceedings of the third {ACM} conference on {Digital} libraries}, May 1998.
\newblock ISBN 978-0-89791-965-4.

\bibitem[Sen et~al.(2008)Sen, Namata, Bilgic, Getoor, Galligher, and Eliassi-Rad]{sen_collective_2008}
Prithviraj Sen, Galileo Namata, Mustafa Bilgic, Lise Getoor, Brian Galligher, and Tina Eliassi-Rad.
\newblock Collective classification in network data.
\newblock \emph{AI magazine}, 29\penalty0 (3):\penalty0 93--93, 2008.

\bibitem[Pei et~al.(2020)Pei, Wei, Chang, Lei, and Yang]{pei2020geom}
Hongbin Pei, Bingzhe Wei, Kevin Chen-Chuan Chang, Yu~Lei, and Bo~Yang.
\newblock Geom-gcn: Geometric graph convolutional networks.
\newblock \emph{arXiv preprint arXiv:2002.05287}, 2020.

\bibitem[Hu et~al.(2020)Hu, Fey, Zitnik, Dong, Ren, Liu, Catasta, and Leskovec]{hu2020open}
Weihua Hu, Matthias Fey, Marinka Zitnik, Yuxiao Dong, Hongyu Ren, Bowen Liu, Michele Catasta, and Jure Leskovec.
\newblock Open graph benchmark: Datasets for machine learning on graphs.
\newblock \emph{Advances in neural information processing systems}, 33:\penalty0 22118--22133, 2020.

\bibitem[Dobson and Doig(2003)]{dobson_distinguishing_2003}
Paul~D. Dobson and Andrew~J. Doig.
\newblock Distinguishing enzyme structures from non-enzymes without alignments.
\newblock \emph{Journal of Molecular Biology}, 330\penalty0 (4):\penalty0 771--783, July 2003.
\newblock ISSN 0022-2836.
\newblock \doi{10.1016/s0022-2836(03)00628-4}.

\bibitem[Wishart~et al.(2018)]{wishart_drugbank_2018}
D~Wishart~et al.
\newblock {DrugBank} 5.0: a major update to the {DrugBank} database for 2018.
\newblock \emph{Nucleic Acids Research}, 46\penalty0 (D1):\penalty0 D1074--D1082, January 2018.
\newblock ISSN 1362-4962.
\newblock \doi{10.1093/nar/gkx1037}.

\bibitem[{National Institutes of Health}(2021)]{nih_aids_screen}
{National Institutes of Health}.
\newblock Aids antiviral screen data.
\newblock \url{https://wiki.nci.nih.gov/spaces/NCIDTPdata/pages/158204006/AIDS+Antiviral+Screen+Data}, 2021.
\newblock Accessed: 2025-05-10.

\bibitem[Jain et~al.(2013)Jain, Ong, Hautier, Chen, Richards, Dacek, Cholia, Gunter, Skinner, Ceder, and Persson]{10.1063/1.4812323}
Anubhav Jain, Shyue~Ping Ong, Geoffroy Hautier, Wei Chen, William~Davidson Richards, Stephen Dacek, Shreyas Cholia, Dan Gunter, David Skinner, Gerbrand Ceder, and Kristin~A. Persson.
\newblock {Commentary: The Materials Project: A materials genome approach to accelerating materials innovation}.
\newblock \emph{APL Materials}, 1\penalty0 (1):\penalty0 011002, 07 2013.
\newblock ISSN 2166-532X.
\newblock \doi{10.1063/1.4812323}.
\newblock URL \url{https://doi.org/10.1063/1.4812323}.

\bibitem[Reijneveld et~al.(2007)Reijneveld, Ponten, Berendse, and Stam]{reijneveld2007application}
Jaap~C Reijneveld, Sophie~C Ponten, Henk~W Berendse, and Cornelis~J Stam.
\newblock The application of graph theoretical analysis to complex networks in the brain.
\newblock \emph{Clinical neurophysiology}, 118\penalty0 (11):\penalty0 2317--2331, 2007.

\bibitem[Boccaletti et~al.(2014)Boccaletti, Bianconi, Criado, {del Genio}, Gómez-Gardeñes, Romance, Sendiña-Nadal, Wang, and Zanin]{boccaletti2014structure}
S.~Boccaletti, G.~Bianconi, R.~Criado, C.I. {del Genio}, J.~Gómez-Gardeñes, M.~Romance, I.~Sendiña-Nadal, Z.~Wang, and M.~Zanin.
\newblock The structure and dynamics of multilayer networks.
\newblock \emph{Physics Reports}, 544\penalty0 (1):\penalty0 1--122, 2014.
\newblock ISSN 0370-1573.
\newblock \doi{https://doi.org/10.1016/j.physrep.2014.07.001}.
\newblock URL \url{https://www.sciencedirect.com/science/article/pii/S0370157314002105}.
\newblock The structure and dynamics of multilayer networks.

\bibitem[Simoes(2011)]{simoes2011agent}
Joana~A Simoes.
\newblock An agent-based/network approach to spatial epidemics.
\newblock \emph{Agent-based models of geographical systems}, 2011.

\bibitem[Holme and Saram{\"a}ki(2012)]{holme2012temporal}
Petter Holme and Jari Saram{\"a}ki.
\newblock Temporal networks.
\newblock \emph{Physics reports}, 519\penalty0 (3):\penalty0 97--125, 2012.

\bibitem[Belbute-Peres et~al.(2020)Belbute-Peres, Economon, and Kolter]{belbute2020combining}
Filipe De~Avila Belbute-Peres, Thomas Economon, and Zico Kolter.
\newblock Combining differentiable pde solvers and graph neural networks for fluid flow prediction.
\newblock In \emph{international conference on machine learning}, pages 2402--2411. PMLR, 2020.

\bibitem[Sanchez-Gonzalez et~al.(2020)Sanchez-Gonzalez, Godwin, Pfaff, Ying, Leskovec, and Battaglia]{sanchez2020learning}
Alvaro Sanchez-Gonzalez, Jonathan Godwin, Tobias Pfaff, Rex Ying, Jure Leskovec, and Peter Battaglia.
\newblock Learning to simulate complex physics with graph networks.
\newblock In \emph{International conference on machine learning}, pages 8459--8468. PMLR, 2020.

\bibitem[Pfaff et~al.(2020)Pfaff, Fortunato, Sanchez-Gonzalez, and Battaglia]{pfaff2020learning}
Tobias Pfaff, Meire Fortunato, Alvaro Sanchez-Gonzalez, and Peter~W Battaglia.
\newblock Learning mesh-based simulation with graph networks.
\newblock \emph{arXiv:2010.03409}, 2020.

\bibitem[Chamberlain et~al.(2021{\natexlab{a}})Chamberlain, Rowbottom, Eynard, Di~Giovanni, Dong, and Bronstein]{chamberlain2021beltrami}
Benjamin Chamberlain, James Rowbottom, Davide Eynard, Francesco Di~Giovanni, Xiaowen Dong, and Michael Bronstein.
\newblock Beltrami flow and neural diffusion on graphs.
\newblock \emph{Advances in Neural Information Processing Systems}, 34:\penalty0 1594--1609, 2021{\natexlab{a}}.

\bibitem[Chamberlain et~al.(2021{\natexlab{b}})Chamberlain, Rowbottom, Gorinova, Bronstein, Webb, and Rossi]{chamberlain2021grand}
Ben Chamberlain, James Rowbottom, Maria~I Gorinova, Michael Bronstein, Stefan Webb, and Emanuele Rossi.
\newblock Grand: Graph neural diffusion.
\newblock In \emph{International Conference on Machine Learning}, 2021{\natexlab{b}}.

\bibitem[Eliasof et~al.(2021)Eliasof, Haber, and Treister]{eliasof2022}
Moshe Eliasof, Eldad Haber, and Eran Treister.
\newblock Pde-gcn: Novel architectures for graph neural networks motivated by partial differential equations.
\newblock In \emph{Neural Information Processing Systems}, 2021.
\newblock URL \url{https://api.semanticscholar.org/CorpusID:236912752}.

\bibitem[Thorpe et~al.(2022)Thorpe, Nguyen, Xia, Strohmer, Bertozzi, Osher, and Wang]{thorpe2022grand++}
Matthew Thorpe, Tan~Minh Nguyen, Heidi Xia, Thomas Strohmer, Andrea Bertozzi, Stanley Osher, and Bao Wang.
\newblock Grand++: Graph neural diffusion with a source term.
\newblock In \emph{International Conference on Learning Representation (ICLR)}, 2022.

\bibitem[Shuman et~al.(2013)Shuman, Narang, Frossard, Ortega, and Vandergheynst]{shuman2013emerging}
David~I Shuman, Sunil~K Narang, Pascal Frossard, Antonio Ortega, and Pierre Vandergheynst.
\newblock The emerging field of signal processing on graphs: Extending high-dimensional data analysis to networks and other irregular domains.
\newblock \emph{IEEE signal processing magazine}, 30\penalty0 (3):\penalty0 83--98, 2013.

\bibitem[Ortega et~al.(2018)Ortega, Frossard, Kova{\v{c}}evi{\'c}, Moura, and Vandergheynst]{ortega2018graph}
Antonio Ortega, Pascal Frossard, Jelena Kova{\v{c}}evi{\'c}, Jos{\'e}~MF Moura, and Pierre Vandergheynst.
\newblock Graph signal processing: Overview, challenges, and applications.
\newblock \emph{Proceedings of the IEEE}, 106\penalty0 (5):\penalty0 808--828, 2018.

\bibitem[Bo et~al.(2021)Bo, Wang, Shi, and Shen]{bo2021beyond}
Deyu Bo, Xiao Wang, Chuan Shi, and Huawei Shen.
\newblock Beyond low-frequency information in graph convolutional networks.
\newblock In \emph{Proceedings of the AAAI conference on artificial intelligence}, volume~35, pages 3950--3957, 2021.

\bibitem[Nt and Maehara(2019)]{nt2019revisiting}
Hoang Nt and Takanori Maehara.
\newblock Revisiting graph neural networks: All we have is low-pass filters.
\newblock \emph{arXiv preprint arXiv:1905.09550}, 2019.

\bibitem[Chew et~al.(2024)Chew, Hirn, Krishnaswamy, Needell, Perlmutter, Steach, Viswanath, and Wu]{chew2022geometric}
Joyce Chew, Matthew Hirn, Smita Krishnaswamy, Deanna Needell, Michael Perlmutter, Holly Steach, Siddharth Viswanath, and Hau-Tieng Wu.
\newblock Geometric scattering on measure spaces.
\newblock \emph{Applied and Computational Harmonic Analysis}, 70:\penalty0 101635, 2024.
\newblock ISSN 1063-5203.
\newblock \doi{https://doi.org/10.1016/j.acha.2024.101635}.
\newblock URL \url{https://www.sciencedirect.com/science/article/pii/S1063520324000125}.

\bibitem[Strogatz(2018)]{strogatz2018nonlinear}
Steven~H Strogatz.
\newblock \emph{Nonlinear dynamics and chaos: with applications to physics, biology, chemistry, and engineering}.
\newblock CRC press, 2018.

\bibitem[Sprott(2008)]{sprott2008chaotic}
JC~Sprott.
\newblock Chaotic dynamics on large networks.
\newblock \emph{Chaos: An Interdisciplinary Journal of Nonlinear Science}, 18\penalty0 (2), 2008.

\bibitem[Arnold and Wihstutz(2006)]{arnold2006lyapunov}
Ludwig Arnold and Volker Wihstutz.
\newblock Lyapunov exponents: a survey.
\newblock In \emph{Lyapunov Exponents: Proceedings of a Workshop held in Bremen, November 12--15, 1984}, pages 1--26. Springer, 2006.

\bibitem[Spielman(2019)]{spielman2019spectral}
Daniel~A Spielman.
\newblock Spectral and algebraic graph theory, 2019.
\newblock \emph{URL http://cs-www. cs. yale. edu/homes/spielman/sagt. Version dated December}, 19, 2019.

\bibitem[Crane et~al.(2017)Crane, Weischedel, and Wardetzky]{Crane2017}
Keenan Crane, Clarisse Weischedel, and Max Wardetzky.
\newblock The heat method for distance computation.
\newblock \emph{Commun. ACM}, 60\penalty0 (11):\penalty0 90--99, October 2017.
\newblock ISSN 0001-0782.
\newblock \doi{10.1145/3131280}.
\newblock URL \url{http://doi.acm.org/10.1145/3131280}.

\bibitem[Münch and Wojciechowski(2019)]{munch2019}
Florentin Münch and Radosław~K. Wojciechowski.
\newblock Ollivier {R}icci curvature for general graph {L}aplacians: Heat equation, {L}aplacian comparison, non-explosion and diameter bounds.
\newblock \emph{Advances in Mathematics}, 356:\penalty0 106759, 2019.
\newblock ISSN 0001-8708.
\newblock \doi{https://doi.org/10.1016/j.aim.2019.106759}.
\newblock URL \url{https://www.sciencedirect.com/science/article/pii/S000187081930369X}.

\bibitem[Gilmer et~al.(2017)Gilmer, Schoenholz, Riley, Vinyals, and Dahl]{gilmer2017neural}
Justin Gilmer, Samuel~S Schoenholz, Patrick~F Riley, Oriol Vinyals, and George~E Dahl.
\newblock Neural message passing for quantum chemistry.
\newblock In \emph{International conference on machine learning}, pages 1263--1272. PMLR, 2017.

\bibitem[Velickovic et~al.(2017)Velickovic, Cucurull, Casanova, Romero, Lio, Bengio, et~al.]{velickovic2017graph}
Petar Velickovic, Guillem Cucurull, Arantxa Casanova, Adriana Romero, Pietro Lio, Yoshua Bengio, et~al.
\newblock Graph attention networks.
\newblock \emph{stat}, 1050\penalty0 (20):\penalty0 10--48550, 2017.

\bibitem[Xu et~al.(2019)Xu, Hu, Leskovec, and Jegelka]{xu2018powerful}
Keyulu Xu, Weihua Hu, Jure Leskovec, and Stefanie Jegelka.
\newblock How powerful are graph neural networks?
\newblock In \emph{7th International Conference on Learning Representations, {ICLR}}, 2019.
\newblock URL \url{https://openreview.net/forum?id=ryGs6iA5Km}.

\bibitem[Ramp{\'a}{\v{s}}ek et~al.(2022)Ramp{\'a}{\v{s}}ek, Galkin, Dwivedi, Luu, Wolf, and Beaini]{rampavsek2022recipe}
Ladislav Ramp{\'a}{\v{s}}ek, Michael Galkin, Vijay~Prakash Dwivedi, Anh~Tuan Luu, Guy Wolf, and Dominique Beaini.
\newblock Recipe for a general, powerful, scalable graph transformer.
\newblock \emph{Advances in Neural Information Processing Systems}, 35:\penalty0 14501--14515, 2022.

\bibitem[Yan et~al.(2022)Yan, Ma, Gao, Tang, Wang, and Chen]{yan_neural_2022}
Zuoyu Yan, Tengfei Ma, Liangcai Gao, Zhi Tang, Yusu Wang, and Chao Chen.
\newblock Neural {Approximation} of {Graph} {Topological} {Features}, November 2022.
\newblock URL \url{http://arxiv.org/abs/2201.12032}.
\newblock arXiv:2201.12032 [cs].

\bibitem[Brandes et~al.(2022)Brandes, Ofer, Peleg, Rappoport, and Linial]{brandes2022proteinbert}
Nadav Brandes, Dan Ofer, Yam Peleg, Nadav Rappoport, and Michal Linial.
\newblock Proteinbert: a universal deep-learning model of protein sequence and function.
\newblock \emph{Bioinformatics}, 38\penalty0 (8):\penalty0 2102--2110, 2022.

\bibitem[Fabian et~al.(2020)Fabian, Edlich, Gaspar, Segler, Meyers, Fiscato, and Ahmed]{fabian2020molecular}
Benedek Fabian, Thomas Edlich, H{\'e}l{\'e}na Gaspar, Marwin Segler, Joshua Meyers, Marco Fiscato, and Mohamed Ahmed.
\newblock Molecular representation learning with language models and domain-relevant auxiliary tasks.
\newblock \emph{arXiv preprint arXiv:2011.13230}, 2020.

\bibitem[Cheng et~al.(2021)Cheng, Zhang, and Dong]{cheng2021geometric}
Jiucheng Cheng, Chunkai Zhang, and Lifeng Dong.
\newblock A geometric-information-enhanced crystal graph network for predicting properties of materials.
\newblock \emph{Communications Materials}, 2\penalty0 (1):\penalty0 92, 2021.

\bibitem[Adams et~al.(2016)Adams, Chepushtanova, Emerson, Hanson, Kirby, Motta, Neville, Peterson, Shipman, and Ziegelmeier]{adams_persistence_2016}
H.~Adams, S.~Chepushtanova, T.~Emerson, E.~Hanson, M.~Kirby, F.~Motta, R.~Neville, C.~Peterson, P.~Shipman, and L.~Ziegelmeier.
\newblock Persistence {Images}: {A} {Stable} {Vector} {Representation} of {Persistent} {Homology}, 2016.
\newblock URL \url{http://arxiv.org/abs/1507.06217}.
\newblock arXiv:1507.06217.

\bibitem[Ollivier(2007)]{ollivier_ricci_2007}
Yann Ollivier.
\newblock {R}icci curvature of {{M}arkov} chains on metric spaces, July 2007.
\newblock URL \url{http://arxiv.org/abs/math/0701886}.
\newblock arXiv:math/0701886.

\bibitem[Cohen-Steiner et~al.(2009)Cohen-Steiner, Edelsbrunner, and Harer]{cohen-steiner_extending_2009}
David Cohen-Steiner, Herbert Edelsbrunner, and John Harer.
\newblock Extending {Persistence} {Using} {Poincaré} and {Lefschetz} {Duality}.
\newblock \emph{Foundations of Computational Mathematics}, 9\penalty0 (1):\penalty0 79--103, February 2009.
\newblock ISSN 1615-3383.
\newblock \doi{10.1007/s10208-008-9027-z}.
\newblock URL \url{https://doi.org/10.1007/s10208-008-9027-z}.

\bibitem[Erdos et~al.(1960)Erdos, R{\'e}nyi, et~al.]{erdos1960evolution}
Paul Erdos, Alfr{\'e}d R{\'e}nyi, et~al.
\newblock On the evolution of random graphs.
\newblock \emph{Publ. math. inst. hung. acad. sci}, 5\penalty0 (1):\penalty0 17--60, 1960.

\bibitem[Zomorodian and Carlsson(2004)]{zomorodian2004computing}
Afra Zomorodian and Gunnar Carlsson.
\newblock Computing persistent homology.
\newblock In \emph{Proceedings of the twentieth annual symposium on Computational geometry}, pages 347--356, 2004.

\bibitem[Donnat et~al.(2018)Donnat, Zitnik, Hallac, and Leskovec]{donnat2018learning}
Claire Donnat, Marinka Zitnik, David Hallac, and Jure Leskovec.
\newblock Learning structural node embeddings via diffusion wavelets.
\newblock In \emph{Proceedings of the 24th ACM SIGKDD international conference on knowledge discovery \& data mining}, pages 1320--1329, 2018.

\bibitem[Kiani et~al.(2024)Kiani, Fesser, and Weber]{kiani2024unitary}
Bobak Kiani, Lukas Fesser, and Melanie Weber.
\newblock Unitary convolutions for learning on graphs and groups.
\newblock \emph{Advances in Neural Information Processing Systems}, 37:\penalty0 136922--136961, 2024.

\bibitem[Defferrard et~al.(2016)Defferrard, Bresson, and Vandergheynst]{defferrard2016convolutional}
Micha{\"e}l Defferrard, Xavier Bresson, and Pierre Vandergheynst.
\newblock Convolutional neural networks on graphs with fast localized spectral filtering.
\newblock \emph{Advances in neural information processing systems}, 2016.

\bibitem[Sprott(1994)]{sprott1994some}
J~Clint Sprott.
\newblock Some simple chaotic flows.
\newblock \emph{Physical review E}, 50\penalty0 (2):\penalty0 R647, 1994.

\bibitem[Sprott(1997)]{sprott1997some}
JC~Sprott.
\newblock Some simple chaotic jerk functions.
\newblock \emph{American Journal of Physics}, 65\penalty0 (6):\penalty0 537--543, 1997.

\bibitem[Coddington(2012)]{coddington2012introduction}
Earl~A Coddington.
\newblock \emph{An introduction to ordinary differential equations}.
\newblock Courier Corporation, 2012.

\bibitem[Fallahgoul et~al.(2017)Fallahgoul, Focardi, and Fabozzi]{FALLAHGOUL201781}
Hasan~A. Fallahgoul, Sergio~M. Focardi, and Frank~J. Fabozzi.
\newblock 7 - continuous-time random walk and fractional calculus.
\newblock In Hasan~A. Fallahgoul, Sergio~M. Focardi, and Frank~J. Fabozzi, editors, \emph{Fractional Calculus and Fractional Processes with Applications to Financial Economics}, pages 81--90. Academic Press, 2017.
\newblock ISBN 978-0-12-804248-9.
\newblock \doi{https://doi.org/10.1016/B978-0-12-804248-9.50007-3}.
\newblock URL \url{https://www.sciencedirect.com/science/article/pii/B9780128042489500073}.

\bibitem[Morris et~al.(2020)Morris, Kriege, Bause, Kersting, Mutzel, and Neumann]{morris2020tudataset}
C.~Morris, N.~Kriege, F.~Bause, K.~Kersting, P.~Mutzel, and M.~Neumann.
\newblock Tudataset: A collection of benchmark datasets for learning with graphs.
\newblock \emph{arXiv:2007.08663}, 2020.

\bibitem[Schomburg et~al.(2004)Schomburg, Chang, Ebeling, Gremse, Heldt, Huhn, and Schomburg]{schomburg_brenda_2004}
Ida Schomburg, Antje Chang, Christian Ebeling, Marion Gremse, Christian Heldt, Gregor Huhn, and Dietmar Schomburg.
\newblock Brenda, the enzyme database: updates and major new developments.
\newblock \emph{Nucleic acids research}, 32\penalty0 (suppl\_1):\penalty0 D431--D433, 2004.

\bibitem[Debnath et~al.(1991)Debnath, Lopez~de Compadre, Debnath, Shusterman, and Hansch]{debnath1991structure}
Asim~Kumar Debnath, Rosa~L Lopez~de Compadre, Gargi Debnath, Alan~J Shusterman, and Corwin Hansch.
\newblock Structure-activity relationship of mutagenic aromatic and heteroaromatic nitro compounds. correlation with molecular orbital energies and hydrophobicity.
\newblock \emph{Journal of medicinal chemistry}, 34\penalty0 (2):\penalty0 786--797, 1991.

\bibitem[Wang et~al.(2020)Wang, Shen, Huang, Wu, Dong, and Kanakia]{10.1162/qss_a_00021}
Kuansan Wang, Zhihong Shen, Chiyuan Huang, Chieh-Han Wu, Yuxiao Dong, and Anshul Kanakia.
\newblock Microsoft academic graph: When experts are not enough.
\newblock \emph{Quantitative Science Studies}, 1\penalty0 (1):\penalty0 396--413, 02 2020.
\newblock ISSN 2641-3337.
\newblock \doi{10.1162/qss_a_00021}.
\newblock URL \url{https://doi.org/10.1162/qss\_a\_00021}.

\bibitem[Kriege and Mutzel(2012)]{kriege_subgraph_2012}
Nils Kriege and Petra Mutzel.
\newblock Subgraph matching kernels for attributed graphs.
\newblock In \emph{Proceedings of the 29th {International} {Conference} on {International} {Conference} on {Machine} {Learning}}, {ICML}'12, 2012.
\newblock ISBN 978-1-4503-1285-1.

\bibitem[Wu et~al.(2017)Wu, Ramsundar, Feinberg, Gomes, Geniesse, Pappu, Leswing, and Pande]{wu_moleculenet_2017}
Zhenqin Wu, Bharath Ramsundar, Evan N. Feinberg, Joseph Gomes, Caleb Geniesse, Aneesh~S. Pappu, Karl Leswing, and Vijay Pande.
\newblock {MoleculeNet}: a benchmark for molecular machine learning †{Electronic} supplementary information ({ESI}) available. {See} {DOI}: 10.1039/c7sc02664a.
\newblock \emph{Chemical Science}, 9\penalty0 (2):\penalty0 513--530, October 2017.
\newblock ISSN 2041-6520.
\newblock \doi{10.1039/c7sc02664a}.
\newblock URL \url{https://www.ncbi.nlm.nih.gov/pmc/articles/PMC5868307/}.

\end{thebibliography}

\clearpage


\appendix

\section{Related work}
\label{sec: related}

There is a long history of studying dynamics on graphs. For example, \citet{reijneveld2007application,boccaletti2014structure,simoes2011agent,holme2012temporal} analyze complex interactions such as brain processes, social networks, and spatial epidemics on graphs. 
The study of graph dynamics has recently crossed into the field of deep learning. with \citet{belbute2020combining,sanchez2020learning,pfaff2020learning} using neural networks to simulate complex phenomena on irregularly structured domains. 
 
Differing from the above mentioned works, 
there have also been several recent papers which aim to use dynamics as a framework for understanding GNNs. 
\citet{chamberlain2021beltrami} and \citet{chamberlain2021grand} take the perspective that message-passing neural networks can be interpreted as the discretizations of diffusion-type (parabolic) partial differential equations on graph domains where each layer corresponds to a discrete time step. 
They then use this insight to design GRAND, a novel GNN, based on encoding the input node features, running a diffusion process for $T$ seconds and finally applying a decoder network. \citet{thorpe2022grand++} builds on this work by extending it to diffusion equations with ``sources'' placed at the labeled nodes, leading to a new network GRAND\texttt{++}. They then provide an analysis of both GRAND and GRAND\texttt{++} and show that they are related to different graph random walks. GRAND is related to a standard graph random walk, whereas GRAND\texttt{++} is related to a dual random walk started at the the labeled data which can avoid the oversmoothing problem. 
We also note \citet{donnat2018learning}, which the graph heat equation to extract structural information around each node (although not in a neural network context) and \citet{eliasof2022} which used insights from hyperbolic and parabolic PDEs on manifolds to design a GNN that does not suffer from oversmoothing 
as well as \citet{kiani2024unitary} which uses convolution using unitary groups to improve GNNs ability to learn long-range dependencies.

Our network method differs from these previous works in several important ways. Most importantly, whereas \citet{chamberlain2021grand} and \citet{chamberlain2021beltrami} primarily focused on PDEs as a framework for understanding the behavior of message passing operations, here we propose to use the dynamics associated to the heat equation as a new form feature aggregation, replacing traditional message passing operations. Additionally, we consider both the heat equation (the prototypical parabolic PDE) and the wave equation (the prototypical  hyperbolic PDE) as well as chaotic dynamics, whereas previous work \citep{chamberlain2021beltrami,chamberlain2021grand,thorpe2022grand++} has primarily focused on parabolic equations. Notably, similar to GRAND\texttt{++}, the wave-equation and the \chaoseqname{}  versions of \modelname{} do not suffer from oversmoothing. However,  the long-term behavior of these equations differs from the diffusion-with-a-source equation used in GRAND\texttt{++} in that they only depend on the geometry of the network and not on locations of the labeled data. (Additionally, since we does not require labeled data as source locations, our method can be easily adapted to unsupervised problems by removing the MLPs.) 

\section{Detailed Background}\label{appx:sec:bg}

This section is a more detailed version of the background on graph signal processing and dynamics provided in Section \ref{sec:bg}. 

\subsection{Graph Signal Processing}\label{appx:bkgrnd:gsp}

In \emph{graph signal processing}, node features are interpreted as signals (functions) defined on the nodes of a graph~\citep{shuman2013emerging,ortega2018graph}. Each signal can then be decomposed into different frequencies defined in terms of the eigendecomposition of the graph Laplacian and an associated Fourier transform.

Formally, we let  $\mathbf{x}:V\rightarrow\mathbb{R}$, denote a function (signal) defined on the vertices of  a weighted, undirected graph $G=(V,E,w)$ with vertices $V=\{v_1,\ldots, v_n\}$. For convenience, we will identify $\mathbf{x}$ with the vector whose $k$-th entry is $\mathbf{x}(v_k)$. Thus, we will write either $\mathbf{x}(v_k)$ or $\mathbf{x}(k)$, depending on context. We may also write $\mathbf{x}(v)$ if we do not wish to emphasize the ordering of the vertices. 
We let $L$  denote  a  graph Laplacian with eigenvectors $\boldsymbol{\nu}_1,\ldots,\boldsymbol{\nu}_n$ and eigenvalues $0=\lambda_1\leq\lambda_2\leq \ldots \leq \lambda_n$, $L\boldsymbol{\nu}_k=\lambda_k\boldsymbol{\nu}_k$. Unless otherwise specified, we will assume that $L$ is either the unnormalized Laplacian $L_{U}=D-A$ or the symmetric normalized Laplacian $L_{\text{sym}}=D^{-1/2}L_\text{U}D^{-1/2}$, where $D$ and $A$ are the weighted degree and adjacency matrices. In these cases,
we may write $L=U\Lambda U^\mathsf{T}$, where $U$ is a matrix with columns $\boldsymbol{\nu}_1,\ldots,\boldsymbol{\nu}_n$ and $\Lambda$ is a diagonal matrix, $\Lambda_{k,k}=\lambda_k$.\footnote{If $L$ is the random walk Laplacian $L_{\text{rw}}=D^{-1}L_U=D^{-1/2}L_{\text{sym}}D^{1/2}$, we may instead obtain an asymmetric eigendecomposition, $L_{\text{rw}}=(D^{-1/2}U)\Lambda (D^{1/2}U)^\mathsf{T}$, where $L_{\text{sym}}=U\Lambda U^\mathsf{T}$.} The graph Fourier transform can be defined as \(\widehat{\mathbf{x}} = U^{\mathsf{T}} \mathbf{x}\) so that $\widehat{\mathbf{x}}(k)=\langle\boldsymbol{\nu}_k,\mathbf{x}\rangle$. Since the $\boldsymbol{\nu}_k$
 form an orthonormal basis, we obtain the Fourier inversion formula $\mathbf{x}=\sum_{k=1}^n \widehat{\mathbf{x}}(k)\boldsymbol{\nu}_k$. 
 The eigenvectors $\boldsymbol{\nu}_k$ are referred to as \emph{Fourier modes} and the eigenvalues $\lambda_k$ are interpreted as \emph{frequencies}. Therefore, the Fourier inversion formula can be thought of as decomposing a signal $\mathbf{x}$ into different the superposition of Fourier modes at different frequencies.
 
Standard message passing neural networks are known to essentially perform low-pass filtering \cite{bo2021beyond,nt2019revisiting}; i.e., they  preserve the portion of the signal corresponding to the first one or two eigenvectors, while suppressing the rest of the signal.
As we discussed in Section \ref{sec: frequency characteristics}, the waveforms utilized in DYMAG may span a broader range of frequency behavior and can act highlight different aspects of the frequency spectrum by acting as either as low-pass, high-pass, or band-pass filters.


\subsection{Heat and Wave Dynamics on a Graph}
\label{appx:bckgrnd:heat_wave}

For $\alpha > 0$, we define the $\alpha$-fractional graph Laplacian by $L^\alpha := U \Lambda ^\alpha U^\mathsf{T}$.
We note that $L^\alpha$ has the same eigenvectors as $L$ and the eigenvalues of $L^\alpha$ are given by $\lambda_k^\alpha$, i.e., $L^\alpha\boldsymbol{\nu}_k=\lambda_k^\alpha\boldsymbol{\nu}_k$. Additionally, we see that when $\alpha=1/m$ for some $m\in\mathbb{N}$, we have $(L^{1/m})^m=L.$ 
For each $i$, we let \(\delta_i\) denote the Dirac signal at $i$ given by
$
\delta_i(k)=1
$ if $i=k$, and $\delta_i(k)=0$ otherwise.
We say that a function $u^{(i)}_H(v,t)$ solves the \emph{$\alpha$-fractional heat equation} with a initial value $\delta_i$ if 
\begin{equation}
\label{appx:eqn:heat_eq_definition}
    -L^\alpha u^{(i)}_H(v,t) = \partial_t u^{(i)}_H(v,t),\quad u^{(i)}_H(v,0)=\delta_i(v).
\end{equation}

We say that $u^{(i)}_W$ solves the \emph{$\alpha$-fractional wave equation} with initial Dirac data 
$\delta_i$ and an initial velocity $c\delta_i$ (where $c$ is a constant) if
\begin{align}
-L^\alpha u^{(i)}_W(v,t) = \partial^2_t u^{(i)}_W(v,t),\quad u^{(i)}_W(v,0)=\delta_i(v), \quad \partial_t u^{(i)}_W(v,0)=c\delta_i(v).\label{appx:eqn:wave_eq_def}
\end{align}

If $G$ is connected, solutions to the heat and wave equations are given explicitly by
\begin{align}
\label{appx:eqn:Heat-app}
u^{(i)}_H(v,t) &= \sum_{k=1}^n e^{-t\lambda_k^\alpha}\langle \boldsymbol{\nu}_k, \delta_i \rangle \boldsymbol{\nu}_k(v), \quad\text{and}\\
u^{(i)}_W(v,t) &=\sum_{k=1}^n \cos(\sqrt{\lambda_k^\alpha} t)\langle \boldsymbol{\nu}_k,\delta_i \rangle \boldsymbol{\nu}_k(v)  + t\langle \boldsymbol{\nu}_1,c\delta_i \rangle\boldsymbol{\nu}_1 
+\sum_{k=2}^n\frac{1}{\sqrt{\lambda_k^\alpha}}\sin(\sqrt{\lambda_k^\alpha} t)\langle \boldsymbol{\nu}_k,c\delta_i\rangle\boldsymbol{\nu}_k(v). \label{appx:eqn:Wave-app}
\end{align}
We note that Eqns \ref{eqn:Heat} and \ref{eqn:Wave} can also be adapted to disconnected graphs with simple modifications.
Furthermore, following Remark 1 in \citet{chew2022geometric}, we note that the solutions $u^{(i)}_H(v,t)$ and $u^{(i)}_W(v,t)$ defined in Eqn. \ref{eqn:Heat} and \ref{eqn:Wave} do not depend on the choice of orthonormal basis for the graph Laplacian, see Section \ref{appendix:joyce_statement} for details.

\subsection{Chaotic Dynamics on a Graph}
\label{appx:bkgrnd:chaos}

We next consider dynamics exhibiting chaos, a behavior that may be informally summarized as ``aperiodic long-term behavior in a deterministic system that exhibits sensitive dependence on initial conditions" \citep{strogatz2018nonlinear}. 
As a prototypical example of chaos on graphs, we consider the general, complex, and nonlinear graph dynamics on graphs described in \citet{sprott2008chaotic}: 

\resizebox{0.9\hsize}{!}{%
\begin{minipage}{\hsize}
\begin{equation}
\smash[t]{\frac{d}{dt} u^{(i)}_S(v_k, t)= -b \cdot u^{(i)}_S(v_k, t)+ \tanh\left( \sum_{v_j \in \mathcal{N}(v_k)} c_{k,j} u^{(i)}(v_j, t)\right),\quad u_S(\cdot,0)=\delta_i,}
\label{appx:eqn:sprott_dynamics}
\end{equation}
\end{minipage}
}

where $b$ is a damping coefficient, and the $c_{k,j}$ represent interactions.  We refer to Eq. \ref{eqn:sprott_dynamics} as the \chaoseqname{} equation and denote its solutions by \chaosmath. When $b>0$, solutions \chaosmath{} remain bounded.
In the case of fully connected graphs, 
with $b = 0.25$,  chaotic dynamics (corresponding to positive Lyapunov exponents, see, e.g., \citet{arnold2006lyapunov}) were observed when a sufficiently large fraction of interactions were neither symmetric ($c_{j,k} = c_{k,j}$) nor anti-symmetric ($c_{j,k} = -c_{k,j}$). Sparsely connected networks also exhibited positive Lyapunov exponents with a value of $b = 0.25$ \citep{sprott2008chaotic}. 

\section{Computational Complexity}\label{appx:complexity}


Here, we discuss the computational complexity of DYMAG and show that it may be scaled to large graphs.

We utilize Chebyshev polynomials to approximate $u^{(i)}_H(v, t)$ and $u^{(i)}_W(v,t)$ as defined in Eqn.~\ref{eqn:Heat} and \ref{eqn:Wave} motivated by the success of Chebyshev polynomials in the approximation of spectral graph wavelets \citep{hammond2011wavelets} and GNNs \citep{defferrard2016convolutional}. This removes the need for eigendecomposition (which can have $\mathcal{O}(n^3)$ computational complexity and $\mathcal{O}(n^2)$ memory). The polynomial approximation has linear complexity for sparse graphs \citep{defferrard2016convolutional}. For example, if $G$ is a $k$-nearest neighbor graph and the order of the polynomial is $m$, then the time complexity for solving the heat/wave equation is $O(kmn)$. 


DYMAG’s runtime complexity is $\mathcal{O}(r|E|FT)$, where $|E|$ is the number of edges, $F$ is the number of features, and $r$ is the degree of the Chebyshev polynomial. 
No closed form solution is available for the Sprott dynamics, so we instead use a fourth-order Runge-Kutta method, which has been successfully applied to chaotic systems \citep{sprott1994some,sprott1997some}. This approach has complexity $\mathcal{O}(g|E|FT)$, where $g$ is the number of solver steps between sampled time steps.

For each of the underlying dynamics, DYMAG exhibits linear complexity, with respect to the number of vertices for sparse graphs (setting $|E|=n\bar{d}$, where $\bar{d}$ is the average degree). This makes it  efficient and scalable to large graphs, where the focus is often on local or node-level properties rather than global topological properties (although predicting global properties is the primary focus of our work). In such cases, local properties can be examined with a smaller feature set of size $F' \ll n$, by concentrating on subgraphs around nodes of interest. Parallelization is feasible for large sparse graphs since the $r$-th order Chebyshev polynomials act over $r$-hop neighborhoods, allowing DYMAG to scale with standard MPNN techniques. Furthermore, the feature space $F$ can be selected or adjusted to be small by utilizing random features, Diracs on a subset of nodes, or natural graph signals.

\section{Proofs of Theoretical Results}\label{app: proofs}

\begin{subsection}{Proof of independence of eigenbasis}
\label{appendix:joyce_statement}
Here we provide a detailed proof for the result that Equations \ref{eqn:Heat} and \ref{eqn:Wave} are invariant to the choice of the Laplacian eigenbasis, as mentioned in \Cref{sec:bckgrnd_heat_wave}.

The solutions $u^{(i)}_H$ and $u^{(i)}_W$ do not depend on the choice of eigenbasis (even when the eigenvalues have multiplicity greater than one). To see this, let $\mathcal{S}_\lambda$ be the set of \emph{distinct} eigenvalues of $L$. For $\lambda\in\mathcal{S}_\lambda,$ let $E_\lambda$ be the corresponding eigenspace, i.e., the linear space spanned by the set of $\boldsymbol{\nu}_k$ such that $\lambda_k=\lambda$. Let $\pi_\lambda$ denote the corresponding projection operator, i.e.,
$$
\pi_\lambda\mathbf{x}=\sum_{k:\lambda_k = \lambda} \langle \boldsymbol{\nu}_k, \mathbf{x} \rangle \boldsymbol{\nu}_k,
$$
for all $\mathbf{x}\in\mathbb{R}^n$, and observe that $\pi_\lambda$ is independent of choice of eigenbasis. Then, from Eqn.~\ref{eqn:Heat}, we may then write,

\begin{align*}
    u^{(i)}_H(v,t) &= \sum_{k =1}^n e^{-t\lambda_k^\alpha} \langle \boldsymbol{\nu}_k, \delta_i \rangle \boldsymbol{\nu}_k(v) \\
    &= \sum_{\lambda \in \mathcal{S}_\lambda} e^{-t\lambda^\alpha} \sum_{k:\lambda_k = \lambda} \langle \boldsymbol{\nu}_k, \delta_i \rangle \boldsymbol{\nu}_k(v) \\
    &= \sum_{\lambda \in \mathcal{S}_\lambda} e^{-t\lambda^\alpha} \pi_\lambda \delta_i(v).
\end{align*}

This establishes that $u^{(i)}_H$ is independent of the choice of basis. The argument for $u^{(i)}_W$ is identical other than using Eqn.~\ref{eqn:Wave} in place of Eqn.~\ref{eqn:Heat}.
\end{subsection}

\begin{subsection}{Proofs of propositions}

Below, we prove \Cref{prop:bandpass}, (restated below) the band-pass properties of DYMAG through the waveforms. 

\textbf{Proposition \ref{prop:bandpass}.\;}
\textit{(Band-pass information)} DYMAG is able to extract band-pass, or even multi-band-pass information information from the node features.

To make Proposition \ref{prop:bandpass} more precise, we will separate it out into two propositions. However, first, we will introduce some notation and definitions.

We define heat-kernel  as $H^t=e^{-tL^{\alpha}}$, i.e.,
$$
H^t=U\operatorname{diag}\left(\exp(-t{\lambda_1})^\alpha,\dots,\exp(-t\lambda_n^\alpha)\right)U^\mathsf{T}.
$$
We observe that we may rewrite the solution to the heat equation, with any initial condition $u_H(\cdot,0)$ as 
\begin{equation}\label{eqn: heat kerenl solution}
u_H(\cdot,t)=\sum_{m=1}^n e^{-t\lambda_m^\alpha}\langle \boldsymbol{\nu}_m, u(\cdot,0) \rangle \boldsymbol{\nu}_m=H^tu_H(\cdot,0).
\end{equation}
In particular, we have $\mathbf{u}_{i,k}=H^{t_k}\delta_i$. 
Thus, we see that the features
\begin{align*}
  h_{j,k}^{(i)}
    &= \langle \mathbf u_{i,k},\mathbf x_j\rangle
\end{align*}
defined in Eqn.~\ref{eq:inner-prod}
can be rewritten as 
\begin{align*}
h^{(i)}_{j,k}&=\sum_{\ell=1}^nH^{t_k}\delta_i(\ell)\mathbf{x}_j(\ell)\\
&=\sum_{\ell=1}^n\sum_{m=1}^n e^{-t\lambda_m^\alpha}\langle \boldsymbol{\nu}_m, \delta_i \rangle \boldsymbol{\nu}_m(\ell)\mathbf{x}_j(\ell)\\
&=\sum_{\ell=1}^n\sum_{m=1}^n e^{-t\lambda_m^\alpha} \boldsymbol{\nu}_m(i) \boldsymbol{\nu}_m(\ell)\mathbf{x}_j(\ell)\\
&=(H^{t_k}\mathbf{x}_j)(i).
\end{align*}

Next, for fixed times $t_1<t_2$, we define   
\begin{equation}\label{eqn: Psi}
\Psi_{t_1,t_2}=H^{t_1}-H^{t_2}
\end{equation}
as the difference of two heat kernels. 
In the spectral domain, we note that we may write
\begin{equation*}
\Psi_{t_1,t_2}\mathbf{x}=\sum_{m=1}^n (e^{-t_1\lambda_m^\alpha}-e^{-t_2\lambda_m^\alpha})\langle \boldsymbol{\nu}_m, \mathbf{x} \rangle \boldsymbol{\nu}_m.
\end{equation*}
The function, \begin{equation}\label{eqn: psi frequency}
  \psi_{t_1,t_2}(\lambda)=e^{-t_1\lambda^{\alpha}}-e^{-t_2\lambda^{\alpha}},
\end{equation}
which is referred to as the frequency response  is zero both at $\lambda=0$, and as $\lambda\rightarrow\infty$. Its support is concentrated in a \emph{frequency band} centered around 
$\lambda^\star
  =\Bigl(\tfrac{\log(t_2/t_1)}{t_2-t_1}\Bigr)^{1/\alpha}.$ Therefore, $\Psi_{t_1,t_2}$ is referred to as a band-pass filter. We summarize this in the following proposition.

\begin{proposition}[Difference of two heat waveforms is band-pass]\label{appx:prop:heat_band}
    Let $t_1<t_2$ and define $\Psi_{t_1,t_2}$ as in Eqn.~\ref{eqn: Psi}. 
Then the function $\mathbf{x}\mapsto\Psi_{t_1,t_2}\mathbf{x}$ performs a band-pass filtering.

\end{proposition}

\begin{proof}
This follows immediately from the frequency response illustrated in Eqn.~\ref{eqn: psi frequency}
 and the subsequent discussion. \end{proof}

Importantly, we observe that DYMAG has the capacity to learn $\Psi_{t_1,t_2}$ through the MLP in Algorithm 2. Therefore, 
Proposition \ref{appx:prop:heat_band} shows that DYMAG, with the heat-equation has the capacity to perform band-pass filtering. This is in contrast to standard message passing networks which are known to effectively perform low-pass filtering (i.e., averaging type operations). 

We next turn our attention to the wave equation, focusing on the case with zero initial velocity (i.e., $c=0$) for the sake of simplicity. Similar to the heat-case (Eqn. \ref{eqn: heat kerenl solution}), we may define a wave kernel by $W^{t}=\cos\left(t\sqrt{L^{\alpha}}\right)=U\operatorname{diag}\left(\cos(t\sqrt{\lambda_1^\alpha}),\dots,\cos(t\sqrt{\lambda_n^\alpha})\right)U^{\mathsf T}$ and observe that the solution to the wave equation, with initial condition $u_W(\cdot,0)$ is given by  $W^{t}u_W(\cdot,0)$.
In the spectral domain, we may write
\begin{equation}
    \Phi_t\mathbf x=\sum_{m=1}^n \cos(t\sqrt{\lambda_m^\alpha})\langle \boldsymbol{\nu}_m, \mathbf{x} \rangle \boldsymbol{\nu}_m.\label{eqn: phi frequency}
\end{equation}
The associated frequency response is given by        \begin{equation}
        \label{eqn: wave frequency response} \psi(\lambda) = \cos(t\sqrt{\lambda^\alpha}).\end{equation} This function
 attains its maximum absolute value at $\lambda=\lambda_{m}:=(m\pi/t)^{2/\alpha},\;m=0,1,2,\dots$ (band-passes), and 
             vanishes at $\lambda=\lambda_{m+\frac12}:=((2m+1)\pi/2t)^{2/\alpha}$ (stop-bands).  
        Hence \(u_{\mathrm{W}}^{(i)}\) alternates between preserving and suppressing successive frequency intervals and thus acts as a \emph{multi-band filter}. We summarize this in the following proposition.

\begin{proposition}[Wave equation is multi-band]\label{appx:prop:wave_band}
        Fix a time $t>0$ and, for simplicity, set the initial velocity to zero ($c=0$).
        Consider the wave kernel
        $
        W^{t}=cos\left(t\sqrt{L^{\alpha}}\right)
        $ and define $\Phi_t$ as in \Cref{eqn: phi frequency}. Then the function $\mathbf x\mapsto\Phi_t\mathbf x$ performs a multi-band-pass filtering.
\end{proposition}

\begin{proof}[Proof of \Cref{appx:prop:wave_band}]
This follows from Eqn.~\ref{eqn: wave frequency response} and the subsequent analysis of the maxima and zeros of $\psi(\lambda)$.
    
\end{proof}

We now turn out attention to Proposition \ref{prop:connected_components informal}.
    



\textbf{Proposition \ref{prop:connected_components informal}.\;} 
\textit{(Identification of Connected Components)}
Let $u^{(i)}(v,t)$ denote the solution to the heat equation, wave equation, or Sprott chaotic dynamics. Suppose that $G$ is not connected. Then, for any $v$ which is not in the same connected component as $v_i$, and all $t\geq0$, we have $u^{(i)}(v,t) = 0.$

\begin{proof}[Proof of Proposition \ref{prop:connected_components informal}]

First observe that it suffices to prove the result  for $0\leq t\leq T$ where $T$ is arbitrary (since we may then let $T\rightarrow \infty$).

Let $u^{(i)}$ be a solution to the differential equation and consider the function $\tilde{u}^{(i)}$ defined by 
$$\tilde{u}^{(i)}(v,t)=\begin{cases}
u^{(i)}(v,t)\quad&\text{if } v\in \mathcal{C}\\
0\quad&\text{otherwise}
\end{cases},$$
where $\mathcal{C}$ is the connected component containing $v_i$.

We observe that $\tilde{u}^{(i)}$ is also a solution since the right hand side of each differential equation is localized in the sense that no energy passes between components
and the initial condition has support contained in $\mathcal{C}$. However, since the right hand side of all three differential equations is Lipschitz on $[0,T]$,  Theorem 5, Section 6 of \citet{coddington2012introduction} implies that there is at most one solution to the differential equation and thus $\tilde{u}^{(i)}=u^{(i)}$ on $V\times [0,T]$. Therefore, since $T$ was arbitrary, we have $u^{(i)}(v,t)=0$ for all $v\notin \mathcal{C}$.

Importantly, we note that the fractional Laplacian $L^\alpha$ is not a graph shift operator, i.e., we may have $L^\alpha_{i,j}\neq 0$ even if $i\neq j$ and $\{v_i,v_j\}\notin E$. However, it is still component-localized in the sense that $L^\alpha_{i,j}\neq 0$ implies that $v_i$ and $v_j$ are in the same connected component. To see this, note that, if $G$ is disconnected, then we can choose an ordering of the vertices so that $L$ is block-diagonal (with the diagonal blocks corresponding to connected components). This implies that we may choose an eigenbasis such that each eigenvector $\boldsymbol{\nu}_i$ has its support (non-zero entries) contained in a single connected component. 
Therefore, writing $L^\alpha$ in its outer-product expansion,
$$
L^\alpha=\sum_{i=1}^n\lambda^\alpha \boldsymbol{\nu}_i\boldsymbol{\nu}_i^\mathsf{T}
$$
implies that $L^\alpha$ is component-localized.
\end{proof}


\textbf{Proposition \ref{prop: full support}.}\; 
\textit{Let $G$ be connected and let $L$ be the random-walk Laplacian $L_{rw}$ (with $\alpha=1)$. Let $u^{(i)}_H(v,t)$ be the solution to the heat equation, Eqn.~ \ref{eqn:Heat}. 
Then $u^{(i)}_H(v,t)>0$ for all $v\in V$ and $t>0$.
}


In order to prove Proposition \ref{prop: full support}, we need the following lemma which relates the heat equation to a continuous time random walker (see e.g., \citet{FALLAHGOUL201781}) $\{X^{\text{continuous}}_t\}_{t\geq 0}$  defined by $X^{\text{continuous}}_t=X^\text{discrete}_{N(t)},$ where $\{X^\text{discrete}_k\}_{k=0}^\infty$ is a standard discrete-time random walker (i.e., a walker who moves to a neighboring vertex at each discrete time step) and $\{N(t)\}_{t\geq 0}$ is an ordinary Poisson process.

\begin{lemma}
\label{thm: RW formal}
Let $u^{(i)}_H(v,t)$ be the solution to the heat equation with $L$ chosen to be the random-walk Laplacian, $L=L_{RW}=I-P$ where $P=D^{-1}A$ and initial condition $\delta_i$.  Then $u^{(i)}_H(\cdot,t)$ is the probability distribution of a continuous-time random walker started at $v_i$ at time $t$.
\end{lemma}

\begin{proof}[Proof of Lemma \ref{thm: RW formal}]

We first note that $L=L_{rw}$ can be written in terms of the symmetric normalized Laplacian $L_s=I_n-D^{-1/2}AD^{-1/2}$ as $L_{rw} = D^{-1/2} L_{s} D^{1/2}$. Therefore, $L_{rw}$ is diagonalizable and may be written as  $L_{rw}=S\Lambda S^{-1}$ where $L_s=U \Lambda U^{-1}$ is a diagonalization of $L_s$ and $S=D^{-1/2}U$. This allows us to write 
$$ P = I-L_{rw} = S(I-\Lambda)S^{-1},$$
which implies that for $k\geq 0$, we have 
$$P^k = S(I-\Lambda)^kS^{-1}.$$

We next note that Eqn.~\ref{eqn:Heat} may be written as $u^{(i)}_H(\cdot,t)= H^t\delta_i$, where $H^t$ is the heat kernel defined by
\begin{equation}\label{eqn: heat_kernel}
    H^t:= e^{-tL} =Se^{-t\Lambda} S^{-1}.
\end{equation}

Therefore, it suffices to show that $H^t$ is the $t$-second transition matrix of the continuous-time random walker $X^{\text{continuous}}_t=X^{\text{discrete}}_{N(t)}$.

By the definition of a Poisson process, $N(t)$ is a Poisson random variable with parameter $t$. Thus, for $k\geq 0,$ $t\geq 0$, we have $$\mathbb{P}(N_t=k)=A_t(k) = t^k e^{-t}/ k!.$$ We next observe that for all $\mu\in\mathbb{R}$ we have
\begin{equation}\label{eqn:poisson}
    e^{-t(1- \mu)} = e^{-t}\sum_{k\geq 0}\frac{(t\mu)^k}{k!} = \sum_{k\geq 0} A_t(k)\mu^k.
\end{equation}
Substituting $\lambda = 1 - \mu$ in Eqn.~\ref{eqn:poisson} links the eigenvalues
of $U^t_H$ and $P$ by 
\begin{align*}
    U^t_H &= Se^{-t\Lambda}S^{-1}= S \sum_{k\geq 0}A_t(k)(I_n - \Lambda)^k S^{-1} \\
    &= \sum_{k\geq 0} A_t(k)P^k.
\end{align*}
This implies that $H^t$ is the $t$-second transition matrix of the continuous-time random walker and thus completes the proof. 
\end{proof}

Now we prove Proposition \ref{prop: full support}.

\begin{proof}[Proof of Proposition \ref{prop: full support}]

Let $v\in V$ be arbitrary. 
As shown in \Cref{thm: RW formal}, $u^{(i)}_H(\cdot,t)$ is the probability distribution of a continuous-time random walk with initial distribution $\delta_i$ at time $t$. Therefore, if $d$ is the length of the shortest path from $v$ to $v_i$, then
\begin{align*}
u_H^{(i)}(v,t)&=\mathbb{P}(X^{\text{continuous}}_t=v|X^{\text{discrete}}_0=v_i)\\&\geq \mathbb{P}(N(t)=d)\mathbb{P}(X^{\text{discrete}}_d=v|X^{\text{discrete}}_0=v_i)>0. \qedhere
\end{align*}

\end{proof}






\textbf{Proposition \ref{prop: heat energy short}.\;}\textit{(Heat energy) Let $G$ be connected, and
let $u^{(i)}_H(v,t)$ be as in Eqn.~\ref{eqn:Heat} and let $u^{(i)}_H(t)=u^{(i)}_H(\cdot,t)$. Then, 
    \begin{equation}
    \label{eqn:heat_energy_bounds}
         e^{-2t \lambda_n^\alpha}  \leq \| u^{(i)}_H(\cdot, t)\|_2^2 \leq |\boldsymbol{\nu}_1(i)|^2 + e^{-2t\lambda_2^\alpha} 
    \end{equation}
    for all $t > 0$.
Furthermore, as $t$ converges to infinity we have 
    \begin{equation}
    \label{eqn:heat_energy_bounds_2}
        \lim_{t \rightarrow \infty} u^{(i)}_H(t) = \langle \boldsymbol{\nu}_1, \delta_i \rangle \boldsymbol{\nu}_1=\boldsymbol{\nu}_1(i)\boldsymbol{\nu}_1.
    \end{equation}}
\begin{proof}[Proof of Proposition \ref{prop: heat energy short}]
From Eqn.~\ref{eqn:Heat}, and the fact that $\{\boldsymbol{\nu}_k\}_{k=1}^n$ is an ONB, we see
\begin{align}
    \|u^{(i)}_H(t) \|_2^2 &= \left\langle \sum_{k=1}^n e^{-t\lambda_k^\alpha}\langle \boldsymbol{\nu}_k,\delta_i\rangle \boldsymbol{\nu}_k, \sum_{k=1}^n e^{-t\lambda_k^\alpha} \langle \boldsymbol{\nu}_k,\delta_i\rangle \boldsymbol{\nu}_k\right\rangle \nonumber\\
    &= \sum_{k = 1}^n e^{-2t\lambda_k^\alpha} |\langle \boldsymbol{\nu}_k, \delta_i \rangle|^2. \label{eqn:heat_ONB}
\end{align}
Thus, since $\lambda_1=0$, upper bound in Eqn.~\ref{eqn:heat_energy_bounds} is obtained by:
    \begin{align*}
        \|u^{(i)}_H(t)\|_2^2 &= \sum_{k = 1}^n e^{-2t\lambda_k^\alpha} |\langle \boldsymbol{\nu}_k, \delta_i \rangle|^2 \\ 
        &= |\langle \boldsymbol{\nu}_1, \delta_i \rangle|^2 + \sum_{k = 2}^n e^{-2t\lambda_k^\alpha} |\langle \boldsymbol{\nu}_k, \delta_i \rangle|^2 \\
        &\leq |\langle \boldsymbol{\nu}_1, \mathbf{x} \rangle|^2 + e^{-2t \lambda_2^\alpha} \sum_{k = 2}^n |\langle \boldsymbol{\nu}_k, \delta_i \rangle|^2 \\
        &\leq |\langle \boldsymbol{\nu}_1, \delta_i \rangle|^2 + e^{-2t\lambda_2^\alpha}.
    \end{align*}

The lower bound in Eqn.~\ref{eqn:heat_energy_bounds} may be obtained by noting
\begin{align*}
    \sum_{k = 1}^n e^{-2t\lambda_k^\alpha} |\langle \boldsymbol{\nu}_k, \delta_i \rangle|^2 &\geq e^{-2t\lambda_n^\alpha} \sum_{k = 1}^n  |\langle \boldsymbol{\nu}_k, \delta_i \rangle|^2 = e^{-2t\lambda_n^\alpha}.
\end{align*}
Eqn.~\ref{eqn:heat_energy_bounds_2} immediately follows Eqn.~\ref{eqn:heat_ONB} and the fact that $0 =\lambda_1< \lambda_2 \leq \ldots \leq \lambda_n$.
\end{proof}


\textbf{Proposition \ref{prop: heat between}.\;}
\textit{(Heat energy between graphs)} Let $G$ and $G'$ be graphs on $n$ vertices with fractional Laplacians $L_G^\alpha$ and $L_{G'}^\alpha$ and let $\delta_i$ and $\delta_{i'}$ be initial conditions for Eqn.~\ref{eqn:heat_eq_definition} on $G$ and $G'$. Assume: 
(i) $L_{G'}^\alpha \succcurlyeq L_G^\alpha$, i.e.,  $\mathbf{v}^\mathsf{T} L_{G'}^\alpha \mathbf{v} \geq \mathbf{v}^\mathsf{T} L_{G}^\alpha \mathbf{v}\:\:\text{for all }\mathbf{v}\in\mathbb{R}^n,$
(ii) We have $| \boldsymbol{\nu}_k'(i)|^2 \leq (1 + \eta_k(t)) |\boldsymbol{\nu}_k(i)|^2$  for all $1\leq k \leq n,$ where we also assume $\eta_k(t) := \exp(2t ((\lambda_k')^\alpha-\lambda_k^\alpha)) - 1 \geq 0.$
Then, with $u_H$ and $u_H'$ defined as in Eqn.~\ref{eqn:Heat},
we have 
$\|(u^{(i)}_H)'(\cdot,t)\|^2_2 \leq \|u^{(i)}_H(\cdot, t)\|_2^2.$

\begin{proof}[Proof of Proposition \ref{prop: heat between}]

By Eqn. \ref{eqn:heat_ONB}, we have    

\begin{align*}
    &\|u^{(i)}_H(\cdot, t)\|_2^2-\|(u^{(i)}_H)'(\cdot,t)\|^2_2 \\
    &\hspace{20pt}= \sum_{k = 1}^n e^{-2t\lambda_k^\alpha} |\langle \boldsymbol{\nu}_k, \delta_i \rangle|^2 - e^{-2t\lambda_k'^\alpha}|\langle \boldsymbol{\nu}_k', \delta_{i'} \rangle|^2 \\
    &\hspace{20pt}\geq \sum_{k = 1}^n \left[ e^{-2t\lambda_k^\alpha} - e^{-2t\lambda_k'^\alpha}(1 + \eta_i(t)) \right] |\langle \boldsymbol{\nu}_k, \delta_i \rangle|^2 \\&\hspace{20pt} =0.\qedhere
\end{align*}

\end{proof}


\textbf{Proposition \ref{prop: wave energy}.\;}
\textit{(Wave energy bounds)} Let $u_W^{(i)}(v,t)$ be the solution to the fractional wave equation Eqn.~\ref{eqn:Wave} with initial conditions $u_W^{(i)}(\cdot,0) = \delta_i$ and $\partial_t u_W^{(i)}(\cdot,0) = 0$. Then, for any time $t \geq 0$, the energy of the waveform satisfies
$|\boldsymbol{\nu}_1(i)|^2 \leq \| u_W^{(i)}(\cdot,t) \|_2^2 \leq  1.$

\begin{proof}[Proof of Proposition \ref{prop: wave energy}]
The proof is similar to the proof of Eqn.~\ref{eqn:heat_energy_bounds}. By the same reasoning as in Eqn. \ref{eqn:heat_ONB}, we have 
\begin{align}
\label{eqn:wave_energy_expression}
    \|u^{(i)}_W(t)\|_2^2  &= \sum_{k = 1}^n \cos^2\left(\sqrt{\lambda_k^\alpha}t\right) |\langle \boldsymbol{\nu}_k,\delta_i\rangle|^2.
\end{align}
Therefore,
\begin{align*}
    \|u^{(i)}_W(t)\|_2^2&= \sum_{k = 1}^n \cos^2(\sqrt{\lambda_k^\alpha}t) |\langle \boldsymbol{\nu}_k,\delta_i\rangle|^2 \leq \sum_{k = 1}^n |\langle \boldsymbol{\nu}_k,\delta_i\rangle|^2 = \|\delta_i\|_2^2=1.
\end{align*}
The lower bound follows by noting that since $\lambda_1=0$ we have:
\begin{align*}
     \|u^{(i)}_W(t)\|_2^2&= \sum_{k = 1}^n \cos^2(\sqrt{\lambda_k^\alpha}t) |\langle \boldsymbol{\nu}_k,\delta_i\rangle|^2 \\
     &\geq \cos^2(\sqrt{\lambda_1^\alpha}t) |\langle \boldsymbol{\nu}_1,\delta_i\rangle|^2 \\
     &= |\langle \boldsymbol{\nu}_1,\delta_i\rangle|^2.\\
     &= |\boldsymbol{\nu}_1(i)|^2.\qquad\qquad\qquad\qquad\qquad\qquad\qquad\qedhere
\end{align*}

\end{proof}


\textbf{Proposition \ref{prop:wave_eigenvalue}.\;}
\textit{(Recovery of eigenspectrum from waveforms)} Let $G$ be connected, and let $u_W^{(i)}(v,t)$ be the solution to the fractional wave equation (Eqn.~\ref{eqn:Wave}), with initial conditions $u_W^{(i)}(\cdot,0) = \delta_i$ and $\partial_t u_W^{(i)}(\cdot,0) = 0$. Then, for any fixed node $v$, the sequence of values $u_W^{(i)}(v, t_1), \dots, u_W^{(i)}(v, t_m)$ obtained from time samples can be used to approximate the full Laplacian eigenspectrum $\{ \lambda_k^\alpha \}_{k=1}^n$ up to arbitrary precision, provided sufficient time resolution.

\begin{proof}
Fix $v$. 
 Since $\mathbf{y}=0,$ we may rewrite Eqn.~\ref{eqn:Wave} as 
\begin{equation*}
f(t)\coloneqq u^{(i)}_W(v,t) = \sum_{k = 1}^n \cos(\sqrt{\lambda_k^\alpha} t) c_k(v)
\end{equation*}
where $c_k(v) = \langle \boldsymbol{\nu}_k,\delta_i\rangle \boldsymbol{\nu}_k(v)$ is a constant with respect to time and depends only on the node position.

Now, let $\epsilon >0$ be a degree of precision and choose $K$ such that $\frac{1}{K}<\epsilon$.
 Approximate
$$
f(t)\approx \tilde{f}(t)\coloneqq \sum_{k=1}^n\cos(a_k t)c_k(v)
$$
where $a_k$ is the multiple of $1/K$ such that $|a_k-\sqrt{\lambda_k^\alpha}|<\epsilon$.
The function $\tilde{f}$ has a finite Fourier expansion and therefore is uniquely characterized by finitely many samples which allows us to recover the $a_k$ and thus approximately recover the $\lambda_k$.

\end{proof}


\textbf{Corollary \ref{cor:cycle_length}.\;}
\textit{(Cycle Length)} The size of cycle graph $C_n$ can be determined from the solution to the fractional wave equation at a single node $v$.

\begin{proof}

Denote $C_n$ the cycle graph with vertices numbered $0, \ldots, n-1$ and edges $(v, v+1)$ modulo $n$. Since $C_n$ is 2-regular, the unnormalized and normalized Laplacian differ only by a constant multiple of 2. Therefore, without loss of generality we will focus on the unnormalized Laplacian. 

It is known that an orthogonal eigenbasis is given by $\{\phi_k\}_{k = 0}^{k = \lfloor n/2 \rfloor } \cup \{\psi_k\}_{k = 1}^{\lceil (n-1)/2 \rceil}$ defined:
\begin{align*}
    \phi_k(v) = \cos \left( \frac{2 \pi kv }{n} \right), \qquad
    \psi_k(v) = \sin \left( \frac{2 \pi k v}{n} \right),
\end{align*}
where the corresponding eigenvalues are given by
\begin{equation}\label{eq:eigen_value_period}
    \lambda_k = 2 - 2 \cos \left( \frac{2 \pi k }{n} \right) = 4 \sin^2 \left( \frac{\pi k }{n} \right)
\end{equation}

Consider the case where the wave equation has an initial condition of $\mathbf{y} = 0$ and the initial signal is given by $\mathbf{x} = \delta_a$. The solution to this equation at a fixed node $v$ is given by:

\begin{align}\label{eq:wave_sol_cycle}
u_W(t) = &\sum_{k = 1}^{\lfloor \frac{n}{2} \rfloor} \cos \left( 2 t \sin \left( \frac{\pi k }{n} \right) \right) \cos( 2 \pi k a /n) \phi_k/ \| \phi_k\|^2  \nonumber \\
&+ \sum_{k=1}^{\lceil (n-1)/2 \rceil} \cos \left( 2 t \sin \left( \frac{\pi k }{n} \right) \right) \sin( 2 \pi k a/n) \psi_k/\|\psi_k\|^2,\\
&+\phi_0/\|\phi_0\|^2
\end{align}
where the third term in Eqn.~\ref{eq:wave_sol_cycle} corresponding to the smallest non-zero eigenvalue is nonzero. By Proposition \ref{prop:wave_eigenvalue}
each of the $\lambda_k$ are uniquely determined by our measurements  and 
$$n = \frac{2 \pi}{\cos ^{-1} \left( \frac{2-\lambda_1}{2} \right)}.$$

Thus $n$ is uniquely determined by our measurements.
\end{proof}

\end{subsection}

\section{Implementation Details}
\label{sec:appendix_implementation}

\subsection{Experimental Computation Resources}
\label{app:hardware}

All experiments were conducted on a high-performance computing server equipped with an Intel\textsuperscript{\textregistered} Xeon\textsuperscript{\textregistered} Gold 6240 CPU (18 cores, 36 threads, 2.60 GHz base frequency) and 730~GB of system RAM. The server is configured with 4 NVIDIA A100 GPUs, each with 40~GB of VRAM, enabling efficient parallel training of deep learning models. The system runs on Red Hat Enterprise Linux 8.8 with CUDA version 11.8 and cuDNN 8.5.0. Experiments were executed using PyTorch 2.0.0 and Python 3.10. Unless otherwise stated, all models were trained using mixed precision to optimize memory usage and throughput.

\subsection{\modelname{} Parameters}
\label{app:params}

Here we describe the architecture and training setup of \modelname{} used to generate the experimental results presented in this paper. \modelname{} supports both node-level and graph-level tasks, and is evaluated on datasets comprising either a single large graph (e.g., citation networks such as PubMed) or collections of small graphs (e.g., synthetic graphs, molecular graphs, and protein structures).

We use a stacked architecture consisting of $L = 3$ DYMAG layers, each simulating $K$ discrete time steps of heat or wave dynamics, where $K$ is selected via grid search specific to each dataset. Between layers, we apply a 3-layer MLP with LeakyReLU activations for node-wise transformation.
For all downstream tasks, we use a 5-layer MLP with LeakyReLU activations as the task-specific head.

For node-level tasks such as node classification in citation graphs, we directly use the hidden node embeddings produced by \modelname{} and feed them into the 5-layer MLP to perform classification or regression. For datasets composed of multiple small graphs (e.g., synthetic Erdos–Renyi graphs, molecular graphs, etc.), we apply mean pooling across the node dimension to obtain a graph-level embedding. This pooled embedding is passed to the same 5-layer MLP for classification or regression.


\subsection{Parameters for Sprott Dynamics}
For Sprott dynamics (\Cref{eqn:sprott_dynamics}), in our experiments, we set $b=0.25$ and the coupling coefficients as $\smash{c_{k,j} \sim \frac{1}{\sqrt{n-1}} \left(2 \cdot \mathrm{Bernoulli}(0.5) - 1\right)}$, which assigns each $c_{k,j}$ a value of $\pm \frac{1}{\sqrt{n-1}}$ with equal probability.


\section{Additional Experiments}
\label{app:more_experiments}

\begin{table*}[!htbp]
\small
\caption{\small
Mean squared error (MSE) for predicting Ollivier–Ricci curvature ($\kappa$) and extended persistence images (EP) on Erd\H{o}s–R'enyi and citation graphs. Results are shown as mean (standard deviation). \modelname{} with heat or wave dynamics outperforms all baselines. A uniform signal was used on graphs without node features. Due to computational cost, Ollivier–Ricci curvature was computed only on a 2,000-node subgraph for PubMed and omitted for OGBN-MAG and OGBN-Papers100M. }
\label{tab:merged_results}
\renewcommand{\arraystretch}{1.1}
\centering
\resizebox{\textwidth}{!}{%
\begin{tabular}{l|ccccccccccc}
\toprule
\textbf{Dataset}
  & \textbf{\modelname\textsubscript{(Heat)}}
  & \textbf{\modelname\textsubscript{(Wave)}}
  & \textbf{\modelname\textsubscript{(Sprott)}}
  & \textbf{MPNN}
  & \textbf{GAT}
  & \textbf{GIN}
  & \textbf{GWT}
  & \textbf{GraphGPS}
  & \textbf{GRAND}
  & \textbf{GRAND\texttt{++}}
  & \textbf{Neural EPD Approx.}
\\
\midrule
\multicolumn{12}{c}{\quad\textbf{Ollivier–Ricci Curvature ($\kappa$)}}\\
\cmidrule(lr){1-12}
ER ($p=0.04$, $n=100$) & \best{1.86e-01} & 1.93e-01 & 7.44e+00 & 3.20e+01 & 2.37e+01 & 5.93e+00 & 3.60e-01 & 1.14e+01 & 1.10e+01 & 1.48e+01 &  \\
                       & (4.01e-03) & (8.78e-03) & (2.73e-01) & (1.28e+00) & (5.45e-01) & (2.37e-01) & (2.96e-02) & (1.51e-01) & (3.28e-01) & (1.80e-01) &  \\
ER ($p=0.06$, $n=100$) & 1.80e-01 & \best{1.76e-01} & 7.39e+00 & 3.19e+01 & 2.13e+01 & 2.06e+00 & 3.63e-01 & 8.58e+00 & 9.26e+00 & 1.61e+01 &  \\
                       & (8.03e-03) & (5.28e-03) & (4.47e-01) & (5.47e-01) & (2.08e+00) & (2.62e-02) & (1.34e-02) & (9.51e-02) & (5.83e-01) & (6.00e-01) &  \\
ER ($p=0.08$, $n=100$) & \best{1.78e-01} & 1.79e-01 & 6.81e+00 & 3.19e+01 & 2.99e+01 & 8.62e-01 & 3.47e-01 & 9.13e+00 & 2.27e+00 & 1.35e+01 &  \\
                       & (8.39e-03) & (2.83e-03) & (2.23e-01) & (1.33e+00) & (2.74e-01) & (9.19e-02) & (2.15e-02) & (3.42e-01) & (2.18e-02) & (3.76e-01) &  \\
ER ($p=0.04$, $n=200$) & 3.63e-01 & \best{3.58e-01} & 1.52e+00 & 1.81e+01 & 6.74e+00 & 7.86e-01 & 7.39e-01 & 3.07e+00 & 5.90e-01 & 2.06e+00 &  \\
                       & (9.09e-03) & (1.47e-02) & (8.83e-02) & (1.01e+00) & (5.36e-01) & (1.93e-02) & (6.13e-02) & (8.83e-02) & (5.74e-02) & (1.36e-02) &  \\
ER ($p=0.06$, $n=200$) & 3.19e-01 & \best{2.63e-01} & 5.28e-01 & 1.74e+01 & 4.39e+00 & 3.39e-01 & 6.86e-01 & 1.61e+00 & 7.38e-01 & 1.82e+00 & N/A\\
                       & (6.32e-02) & (1.03e-02) & (1.47e-02) & (1.88e-01) & (1.76e-01) & (1.68e-02) & (8.37e-03) & (7.56e-02) & (9.22e-03) & (2.54e-02) &      \\
ER ($p=0.08$, $n=200$) & \best{2.14e-01} & 2.57e-01 & 5.73e-01 & 1.55e+01 & 6.18e+00 & 4.27e-01 & 5.93e-01 & 9.92e-01 & 4.33e-01 & 1.76e+00 &      \\
                       & (9.55e-03) & (5.22e-02) & (1.35e-02) & (7.13e-01) & (1.41e-01) & (2.64e-02) & (6.76e-03) & (3.25e-02) & (1.28e-02) & (5.41e-02) &      \\
Cora                   & \best{1.73e-02} & 1.85e-02 & 1.34e-01 & 2.40e-01 & 7.36e-01 & 1.56e-01 & 4.07e-02 & 4.41e-01 & 6.81e-02 & 1.74e-01 &      \\
                       & (4.44e-04) & (2.99e-03) & (5.98e-03) & (8.12e-03) & (3.65e-02) & (2.05e-03) & (5.47e-04) & (2.16e-02) & (1.02e-03) & (5.03e-03) &      \\
Citeseer              & \best{2.09e-02} & 3.41e-02 & 1.46e-01 & 2.20e-01 & 9.04e-01 & 1.72e-01 & 3.58e-02 & 2.82e-01 & 1.24e-01 & 3.09e-01 &      \\
                       & (1.04e-03) & (2.29e-02) & (5.76e-03) & (2.82e-03) & (8.06e-03) & (1.51e-03) & (6.59e-04) & (2.63e-02) & (5.66e-03) & (5.61e-03) &      \\
PubMed                 & 7.30e-03 & \best{6.51e-03} & 2.97e-02 & 1.69e+00 & 1.55e-01 & 8.34e-03 & 9.15e-03 & 7.68e-01 & 2.73e-02 & 8.72e-02 &      \\
                       & (3.00e-05) & (1.68e-04) & (2.67e-03) & (9.92e-02) & (7.06e-03) & (5.55e-04) & (5.12e-04) & (1.38e-02) & (6.70e-04) & (1.11e-03) &      \\
\midrule
\multicolumn{12}{c}{\quad\textbf{Extended Persistence Image (EP)}}\\
\cmidrule(lr){1-12}
ER ($p=0.04$, $n=100$) & 1.48e-02 & \best{6.37e-03} & 6.63e-01 & 7.83e-01 & 3.82e+00 & 4.83e-01 & 3.76e-02 & 9.17e-01 & 7.72e-01 & 2.39e+00 & 3.07e-03\\
                       & (1.26e-03) & (3.30e-03) & (1.01e-02) & (1.25e-02) & (1.41e-01) & (1.90e-02) & (3.46e-03) & (4.13e-02) & (4.33e-02) & (1.08e-01) & (5.43e-05)\\
ER ($p=0.06$, $n=100$) & 8.65e-03 & \best{2.79e-03} & 6.24e-01 & 7.35e-01 & 1.57e+00 & 4.09e-01 & 3.54e-02 & 6.31e-01 & 5.72e-01 & 1.15e+00 & 1.37e-03\\
                       & (7.91e-04) & (1.42e-03) & (7.97e-03) & (1.46e-02) & (7.12e-02) & (2.64e-02) & (2.64e-03) & (5.82e-02) & (3.03e-02) & (5.05e-02) & (2.26e-05)\\
ER ($p=0.08$, $n=100$) & 8.82e-03 & \best{2.54e-03} & 5.71e-01 & 9.46e-01 & 6.65e-01 & 3.92e-02 & 3.29e-02 & 4.85e-01 & 1.07e-02 & 8.59e-01 & 1.90e-03\\
                       & (2.59e-04) & (6.41e-04) & (1.57e-02) & (5.08e-02) & (6.56e-02) & (1.86e-03) & (2.03e-03) & (7.19e-03) & (1.14e-04) & (5.42e-02) & (4.81e-05)\\
ER ($p=0.04$, $n=200$) & 8.91e-03 & \best{5.18e-03} & 8.04e-02 & 6.73e-01 & 1.93e-01 & 3.72e-02 & 2.88e-02 & 3.92e-01 & 7.31e-02 & 2.85e-01 & 3.24e-03\\
                       & (9.43e-05) & (1.94e-03) & (2.70e-03) & (2.66e-02) & (2.18e-03) & (2.92e-03) & (1.54e-03) & (2.58e-02) & (2.31e-03) & (1.84e-02) & (1.28e-05)\\
ER ($p=0.06$, $n=200$) & 7.41e-03 & \best{4.76e-03} & 1.39e-01 & 7.29e-01 & 5.48e-01 & 3.69e-02 & 2.57e-02 & 3.68e-01 & 8.39e-02 & 3.28e-01 & 4.30e-03\\
                       & (1.23e-04) & (4.62e-04) & (4.08e-03) & (2.10e-02) & (2.81e-02) & (5.56e-04) & (8.74e-04) & (5.41e-03) & (4.93e-03) & (2.78e-02) & (5.53e-05)\\
ER ($p=0.08$, $n=200$) & \best{4.57e-03} & 1.62e-03 & 1.35e-01 & 1.28e+00 & 1.87e-01 & 3.96e-02 & 2.43e-02 & 3.12e-01 & 4.12e-02 & 3.48e-01 & 1.12e-03\\
                       & (6.46e-05) & (5.01e-04) & (3.22e-03) & (9.03e-02) & (1.52e-02) & (9.58e-04) & (5.02e-04) & (1.15e-02) & (6.35e-04) & (4.65e-03) & (1.73e-05)\\
Cora                   & 1.45e-04 & \best{7.41e-05} & 3.51e-03 & 2.72e-03 & 5.04e-02 & 5.53e-03 & 6.79e-04 & 2.71e-02 & 8.13e-04 & 8.16e-03 & 5.80e-05\\
                       & (9.14e-06) & (1.61e-05) & (3.30e-04) & (1.17e-04) & (3.90e-03) & (1.73e-04) & (1.53e-05) & (1.40e-03) & (2.59e-05) & (1.37e-04) & (3.22e-07)\\
Citeseer              & 3.94e-04 & \best{1.58e-04} & 7.64e-03 & 6.83e-03 & 2.93e-02 & 3.94e-03 & 9.12e-04 & 3.16e-02 & 6.87e-04 & 4.21e-03 & 1.29e-04\\
                       & (2.45e-05) & (2.91e-05) & (6.28e-04) & (6.02e-04) & (1.00e-03) & (7.05e-05) & (4.80e-05) & (7.36e-04) & (2.17e-05) & (7.09e-05) & (9.21e-07)\\
PubMed                 & 7.93e-03 & \best{3.29e-03} & 8.96e-02 & 4.29e-01 & 4.79e-01 & 1.27e-01 & 2.16e-02 & 1.94e-01 & 3.37e-02 & 3.07e-01 & 2.74e-03\\
                       & (3.55e-04) & (5.51e-04) & (5.26e-03) & (2.60e-02) & (3.41e-02) & (1.13e-02) & (2.77e-04) & (1.58e-02) & (7.31e-04) & (2.06e-02) & (6.85e-05)\\
ogbn-mag               & 5.39e-02 & \best{8.01e-03} & --      & 2.93e-01 & 4.47e-01 & 2.49e-01 & 5.83e-02 & 3.78e-01 & --      & --      & 6.73e-03\\
                       & (5.95e-04) & (1.28e-03) &        & (5.77e-03) & (1.89e-02) & (3.04e-03) & (8.10e-04) & (1.24e-02) &        &        & (1.04e-04)\\
ogbn-papers100M        & 9.03e-02 & \best{3.25e-02} & --      & 6.19e-01 & 8.03e-01 & 6.43e-01 & 2.71e-01 & 4.05e-01 & --      & --      & 3.18e-02\\
                       & (6.49e-03) & (7.21e-04) &        & (2.36e-02) & (4.13e-02) & (1.15e-02) & (2.15e-02) & (1.12e-02) &        &        & (1.05e-04)\\
\bottomrule
\end{tabular}%
}
\end{table*}

We present further experimental validation of \modelname{}, focusing on its ability to predict both biomolecular properties and topological descriptors such as curvature.

\begin{table*}[!htbp]
\small
\caption{Accuracy (mean $\pm$ standard deviation) of each model on the ENZYMES, PROTEINS, and MUTAG datasets, averaged over 10-fold cross-validation. DYMAG variants based on heat and wave dynamics achieve the best or second-best performance across all datasets.}
\label{tab:classification}
\centering
\begin{tabular}{lccc}
\toprule
\textbf{Model} & \textbf{ENZYMES} & \textbf{PROTEINS} & \textbf{MUTAG} \\
\midrule
DYMAG\textsubscript{(Heat)}      & \best{$0.82 \pm 0.02$} & $0.54 \pm 0.04$  & $0.79 \pm 0.02$ \\
DYMAG\textsubscript{(Wave)}      & $0.79 \pm 0.02$ & \best{$0.71 \pm 0.02$}  & \best{$0.83 \pm 0.02$} \\
DYMAG\textsubscript{(Sprott)}    & $0.60 \pm 0.04$ & $0.64 \pm 0.03$ & $0.74 \pm 0.03$                \\
MPNN   & $0.63 \pm 0.01$ & $0.67 \pm 0.01$  & $0.76 \pm 0.02$ \\
GAT    & $0.65 \pm 0.01$ & $0.64 \pm 0.01$  & $0.75 \pm 0.02$ \\
GIN    & $0.68 \pm 0.01$ & $0.69 \pm 0.02$  & $0.77 \pm 0.02$ \\
GWT    & $0.66 \pm 0.01$ & $0.66 \pm 0.02$  & $0.72 \pm 0.02$ \\
GraphGPS   & $0.70 \pm 0.02$ & $0.68 \pm 0.02$  & $0.78 \pm 0.02$ \\
GRAND      & $0.71 \pm 0.02$ & $0.65 \pm 0.02$  & $0.76 \pm 0.02$ \\
GRAND\texttt{++}  & $0.74 \pm 0.01$ & $0.69 \pm 0.01$  & $0.80 \pm 0.02$ \\
\bottomrule
\end{tabular}
\end{table*}

In \Cref{tab:merged_results}, we present the complete results for predicting Ollivier–Ricci curvature and extended persistence images using synthetic and real-world datasets comprising of Erd\H{o}s–R\'enyi graphs and citation networks. For all experiments, \modelname{} is trained using a uniform signal as input. Ground truth Ollivier–Ricci curvature values are computed directly from the adjacency matrix, and persistence images are generated using node degree as the filtration function. Results are reported as mean with standard deviation in parentheses. Across all settings, \modelname{} variants with heat or wave dynamics consistently outperform baseline methods. Given the high computational cost of computing Ollivier–Ricci curvature on large graphs, we restricted the PubMed evaluation to a subgraph comprising the 2,000 most highly cited papers and omitted this evaluation for OGBN-MAG and OGBN-Papers100M.

In Table \ref{tab:classification}, we evaluated the performance of \modelname{} on three publicly available datasets for biomolecular graph classification from the TUDatasets benchmark \citep{morris2020tudataset}. The ENZYMES dataset \citep{schomburg_brenda_2004} consists of protein secondary structures with ground truth annotations of catalytic activity. In the PROTEINS dataset \citep{dobson_distinguishing_2003}, the task is to classify whether a protein functions as an enzyme. The MUTAG dataset \citep{debnath1991structure}, contains small nitroaromatic compounds, and the task is to classify their mutagenicity on the \textit{S.typhimurium} bacterium. \modelname{} achieves strong validation accuracy across all datasets, with the wave equation variant performing best on PROTEINS and MUTAG.

In \Cref{tab:results-graph-params}, we present results corresponding to Figure~\ref{fig:bar}, where the task is to recover generating parameters of random graphs. On both Erd\H{o}s–Rényi and stochastic block model (SBM) datasets, \modelname{} achieves the best overall performance, further validating its capacity to capture latent structural properties. 

\subsection{Ollivier-Ricci Curvature}
\label{app:OR_results}

In Riemannian geometry, Ricci curvature quantifies how a space deviates from being locally Euclidean by measuring volume distortion along geodesics. The discrete Ollivier-Ricci formulation extends this notion to graphs by capturing how neighborhood structures contract or expand under local optimal transport, thereby providing a principled measure of local geometric distortion.

To evaluate the capacity of \modelname{} to capture such local geometric properties, we consider the task of predicting node-level Ollivier-Ricci curvature~\citep{ollivier_ricci_2007}. Ground truth curvature values are computed using the \texttt{GraphRicciCurvature} package (v0.5.3.1). The method first calculates edge-level curvature scores via optimal transport between neighborhood distributions, and then aggregates these values to the node level by averaging over all incident edges.

Because Ollivier-Ricci curvature is determined entirely by the graph topology - specifically, the adjacency structure and any edge weights - it does not require node features. We therefore compute curvature values directly from the graph’s adjacency matrix. The resulting node-level values are used as regression targets in a node-level prediction task.

We report results on both real-world and synthetic graphs in \Cref{tab:merged_results}. Due to the computational cost of Ricci curvature estimation on large graphs, we restrict evaluation on PubMed (19,717 nodes) to a subgraph comprising the 2,000 most highly cited nodes.

\subsection{Extended Persistence Image}
\label{appendix: extended persistance image full}

To compute extended persistence diagrams for graphs, we define a scalar filtration function $f: V \rightarrow \mathbb{R}$ that assigns a real value to each node. By default, we use node degree, but the framework supports any scalar-valued function (e.g., centrality, clustering coefficient, or domain-specific metadata).

From this node-level function, edge values are induced by setting $f(u, v) = \max\{f(u), f(v)\}$. We then construct a filtration over the graph using both sublevel and superlevel sets: in the sublevel filtration, nodes and edges are added in order of increasing $f$, capturing the evolution of connected components and cycles; in the superlevel filtration, nodes and edges are included in decreasing order, allowing for the identification of global topological features that persist across the entire graph. The extended persistence diagram combines information from ordinary, relative, and extended homology classes to characterize these multiscale topological changes.

Each extended persistence diagram contains a collection of birth–death pairs for two types of features: dimension 0 features correspond to connected components, while dimension 1 features correspond to cycles. To convert these diagrams into a vector representation, we map each birth–death pair $(b, d)$ to a birth–persistence pair $(b, p)$ where $p = d - b$. We then place a Gaussian kernel (with bandwidth $\sigma = 0.005$) centered at each $(b, p)$ coordinate and discretize the resulting function onto a grid to obtain persistence images~\citep{adams_persistence_2016}. To reflect the distinct statistical profiles of the two types of features, we use different grid resolutions: for dimension 0 features, which are typically short-lived, we use a compact $25 \times 1$ grid; for dimension 1 features, which exhibit greater variability in both birth and persistence, we use a full $25 \times 25$ grid. These two images are flattened and concatenated into a 650-dimensional vector.

We treat persistence image prediction as a graph-level regression task. The pooled graph embedding from \modelname{} is passed through a 5-layer MLP to predict the flattened persistence image vector.

Results using node degree as the filtration function are reported in \Cref{tab:merged_results} (bottom). In \Cref{tab:appendix_clustering_coeff_epi}, we present results obtained using clustering coefficient as the filtration function. The clustering coefficient of a node $v$ is defined as 
$$c(v) = \frac{2T(v)}{\text{deg}(v)(\text{deg}(v)-1)}$$ 
where $T(v)$ is the number of triangles through node $v$ and $\text{deg}(v)$ is the degree of node $v$. Compared to degree-based filtrations, the higher MSE values in this setting suggest that persistence images generated using clustering coefficients are more challenging to predict.

\subsection{Node Classification Accuracy on Homophilic and Heterophilic Datasets}
\label{appendix:node-classification}

In Table \ref{tab:results-node-classification}, we also consider node classification on homophilic (Pubmed, Citeseer, and Cora) and heterophilic (Texas, Wisconsin, and Cornell) networks from \citet{pei2020geom}. On the homophilic real-world graph datasets, we see that the wave version of \modelname{} outperforms the heat/Sprott versions, achieving performance that is roughly comparable with the more standard GNNs. However, on the heterophilic datasets, we see that \modelname{} with chaotic Sprott dynamics outperforms other models.

\vfill

\subsection{Fractional Heat Equation Dynamics}
\label{appendix:fractional}

In Tables \ref{tab:results-epd-fractional} and \ref{tab:results-mols-fractional}, we conduct experiments investigating the role of $\alpha$ in the fractional Laplacian $L^\alpha$. We highlight the case $\alpha=1$ in gray to emphasize that when $\alpha=1$, the fractional Laplacian reduces to the standard (non-fractional) graph Laplacian. The values of $\alpha$ that achieve the best performance are highlighted in blue. In cases where there is a tie and $\alpha=1$ is one of the co-best methods (e.g., in the MP dataset), we highlight the $\alpha=1$ case in gray and the other top-performing method in blue.

Table \ref{tab:results-epd-fractional}, reports the mean squared error (MSE) for predicting extended persistence images. We observe that varying $\alpha$ significantly impacts the model's performance. For most Erdős–Rényi (ER) graphs with $n=100$ nodes, lower values of $\alpha$ yield better performance than the standard Laplacian ($\alpha=1$), suggesting that fractional heat diffusion processes capture relevant graph features more effectively in this context.  For larger graphs with $n=200$ nodes, the optimal $\alpha$ is still lower than 1, though not as low as 0.25.

In Table \ref{tab:results-mols-fractional}, which reports the $R^2$ scores for predicting various geometric and graph topological properties of molecules, we observe that the performance across different $\alpha$ values is relatively similar, indicating robustness to the choice of $\alpha$. For example, on the PROTEINS dataset for predicting dihedral angles, the highest $R^2$ score is $0.89$ at both $\alpha=0.25$ and $\alpha=0.50$, while at $\alpha=1$, the score is $0.87$. In the case of the Materials Project (MP) dataset for predicting band gap, there is a tie in performance between $\alpha=0.50$ and $\alpha=1.00$, both achieving an $R^2$ score of $0.59$. 

Overall, these results demonstrate that non-local smoothing achieved with the fractional Laplacian featuring various $\alpha$ parameters   allows the model to perform better on certain tasks. Specifically, fractional Laplacians with $\alpha < 1$ can enhance performance in recovering the topology of randomly generated graphs, while different values of $\alpha$ do not significantly impact DYMAG's performance on molecular and material science datasets.

\begin{table*}[!htbp]
\renewcommand{\arraystretch}{1.1}
\small
\caption{Mean squared error (MSE) for predicting extended persistence images using vertex degree as the filtration function (lower is better). We compare DYMAG models with wave dynamics across different fractional orders $\alpha$. The first group of ER graphs are generated with $n=100$ nodes, and the second with $n=200$ nodes.}
\centering
\smallskip
\resizebox{\textwidth}{!}{
\begin{tabular*}{1.2\textwidth}{@{\extracolsep{\fill}}c|cccc}
\toprule
\multirow{2}{*}{\textbf{Graph}} & \multicolumn{4}{c}{\textbf{Fraction $\alpha$}} \\ 
                                & 0.25 & 0.50 & 0.75 & 1.00 \\
\midrule
$\text{ER}(p=0.04, n=100)$      & 4.47e-2 $\pm$ 1.0e-3 & 3.51e-2 $\pm$ 7.2e-4 & \best{1.39e-2 $\pm$ 2.6e-4} & 1.48e-02 $\pm$ 1.26e-03\\
$\text{ER}(p=0.06, n=100)$      & \best{3.60e-3 $\pm$ 1.6e-4} & 6.03e-3 $\pm$ 8.3e-5 & 6.24e-3 $\pm$ 1.8e-4 & 8.65e-03 $\pm$ 7.91e-04 \\
$\text{ER}(p=0.08, n=100)$      & \best{4.72e-3 $\pm$ 1.8e-4} & 7.39e-3 $\pm$ 2.1e-4 & 8.60e-3 $\pm$ 3.1e-4 & 8.82e-03 $\pm$ 2.59e-04\\
\midrule
$\text{ER}(p=0.04, n=200)$      & 8.12e-3 $\pm$ 5.0e-4 & \best{7.73e-3 $\pm$ 3.4e-4} & 9.28e-3 $\pm$ 8.5e-4 & 8.91e-03 $\pm$ 9.43e-05\\
$\text{ER}(p=0.06, n=200)$      & 7.85e-3 $\pm$ 3.6e-4 & 4.98e-3 $\pm$ 9.3e-5 & \best{4.52e-3 $\pm$ 1.8e-4} & 7.41e-03 $\pm$ 1.23e-04\\
$\text{ER}(p=0.08, n=200)$      & 5.02e-3 $\pm$ 1.6e-4 & 3.47e-3 $\pm$ 9.7e-5 & \best{1.12e-3 $\pm$ 2.5e-4} & 4.57e-03 $\pm$ 6.46e-05\\
\midrule
Cora                            & 2.84e-3 $\pm$ 3.2e-4 & 1.79e-3 $\pm$ 4.0e-4 & 5.16e-4 $\pm$ 8.0e-6 & \best{1.45e-04 $\pm$ 9.14e-06}\\
Citeseer                        & 1.35e-3 $\pm$ 9.7e-5 & 1.04e-3 $\pm$ 1.0e-4 & 6.34e-4 $\pm$ 1.2e-5 & \best{3.94e-04 $\pm$ 2.45e-05}\\
PubMed                          & 5.62e-3 $\pm$ 8.8e-5 & \best{1.17e-4 $\pm$ 1.0e-5} & 2.35e-4 $\pm$ 7.4e-6 & 7.93e-03 $\pm$ 3.55e-04\\
\bottomrule
\end{tabular*}}
\label{tab:results-epd-fractional}
\end{table*}

\begin{table*}[!htbp]
\small
\caption{\small
Performance of \modelname{} (heat dynamics) across different fractional orders $\alpha$ on four datasets: PROTEINS, DrugBank, Materials Project (MP), and the DTS AIDS Antiviral Screen. We report R$^2$ score (higher is better) for the first three datasets and balanced accuracy for the Antiviral Screen. Results are reported as mean $\pm$ standard deviation over 10‑fold cross-validation.}
\label{tab:results-mols-fractional}
\renewcommand{\arraystretch}{1.1}
\centering
\smallskip
\resizebox{\textwidth}{!}{%
\begin{tabular*}{\textwidth}{@{\extracolsep{\fill}}c|c|cc|c|c}
\toprule
\multirow{2}{*}{$\alpha$}
  & \textbf{PROTEINS}
  & \multicolumn{2}{c|}{\textbf{DrugBank}}
  & \textbf{MP}
  & \textbf{Antiviral Screen} \\
  & Dihedral Angles
  & TPSA
  & \# Aromatic Rings
  & Band Gap
  & Active/Inactive \\
\midrule
0.25
  & \best{$0.89 \pm 0.04$}
  & $0.94 \pm 0.02$
  & \best{$0.96 \pm 0.03$}
  & $0.57 \pm 0.04$
  & $0.52 \pm 0.02$ \\

0.50
  & \best{$0.89 \pm 0.03$}
  & $0.92 \pm 0.02$
  & \best{$0.96 \pm 0.02$}
  & \best{$0.59 \pm 0.01$}
  & \best{$0.56 \pm 0,01$} \\

0.75
  & $0.86 \pm 0.02$
  & $0.92 \pm 0.03$
  & \best{$0.97 \pm 0.03$}
  & $0.58 \pm 0.02$
  & \best{$0.56 \pm 0.02$} \\

1.00
  & \best{$0.89 \pm 0.01$}
  & \best{$0.97 \pm 0.01$}
  & \best{$0.97 \pm 0.02$}
  & \best{$0.61 \pm 0.03$}
  & \best{$0.54 \pm 0.02$} \\
\bottomrule
\end{tabular*}%
}
\end{table*}

\begin{table*}[!htbp]
\renewcommand{\arraystretch}{1.1}
\small
\caption{Mean squared error (MSE) for the prediction of generating parameters of random graphs (lower is better). The number of nodes for each type of random graph is specified in each data column (i.e. $n \in \{100, 250, 500, 1000, 2500\}$).}
\label{tab:results-graph-params}
\centering
\smallskip
\resizebox{\textwidth}{!}{%
\begin{tabular*}{1.2\textwidth}{@{\extracolsep{\fill}}c|ccccc|ccccc}
\toprule
\multirow{2}*{\textbf{Method}} 
  & \multicolumn{5}{c|}{\textbf{Erd\H{o}s–R\'enyi}} 
  & \multicolumn{5}{c}{\textbf{Stochastic Block Model}}\\
  & $100$ & $250$ & $500$ & $1000$ & $2500$ 
  & $100$ & $250$ & $500$ & $1000$ & $2500$ \\
\midrule
\modelname\textsubscript{(Heat)}   
  & \best{7.46e-3} & 7.13e-3  & 3.60e-3  & 4.19e-3  & 3.04e-3  
  & \best{6.41e-1} & \best{8.10e-1} & 1.79 & 4.52 & 11.63 \\

\modelname\textsubscript{(Wave)}   
  & 8.29e-3 & \best{6.58e-3} & \best{3.17e-3} & \best{3.25e-3} & \best{1.04e-3}  
  & 8.25e-1 & 9.40e-1 & \best{1.26} & \best{2.28} & \best{2.35} \\

\modelname\textsubscript{(Sprott)} 
  & 4.33e-2 & 4.92e-2 & 7.08e-3 & 3.68e-3 & 5.49e-3
  & 5.17 & 3.37 & 4.25 & 4.08 & 6.27 \\
  
MPNN                                 
  & 1.37e-2 & 1.14e-2 & 9.26e-3 & 9.49e-3 & 8.02e-3
  & 2.93 & 3.07 & 3.68 & 7.14 & 10.26 \\

GAT                                  
  & 3.05e-2 & 5.60e-2 & 1.35e-2 & 3.74e-2 & 2.69e-2
  & 11.79 & 9.42 & 10.83 & 13.62 & 18.60 \\

GIN                                  
  & 1.08e-2 & 9.37e-3 & 7.74e-3 & 6.98e-3 & 4.81e-3
  & 1.74 & 2.59 & 2.92 & 4.37 & 9.15 \\
  
GWT                                  
  & 9.72e-3 & 1.04e-2 & 6.29e-3 & 6.56e-3 & 5.41e-3        
  & 2.47    & 3.18    & 2.14    & 4.87    & 6.52    \\

GraphGPS                              
  & 5.28e-2 & 8.48e-2  & 1.26e-2 & 1.31e-2 & 8.24e-3       
  & 12.06   & 8.21     & 9.44    & 11.63   & 12.67  \\

GRAND                                
  & 6.36e-2 & 4.22e-2 & 9.27e-3 & 6.58e-3 & 5.30e-3
  & 14.52 & 16.78 & 13.50 & 11.28 & 8.58 \\

GRAND\texttt{++}                     
  & 8.52e-2 & 6.91e-2 & 2.84e-2 & 1.29e-2 & 8.72e-3
  & 23.71 & 26.84 & 19.64 & 16.97 & 15.42 \\
\bottomrule
\end{tabular*}%
}
\end{table*}

\begin{table*}[!htbp]
\renewcommand{\arraystretch}{1.1}
\small
\caption{Node classification accuracy (\%) on homophilic and heterophilic datasets.}
\label{tab:results-node-classification}
\centering
\smallskip 
\resizebox{\textwidth}{!}{%
\begin{tabular*}{1.2\textwidth}{@{\extracolsep{\fill}}c|ccc|ccc}
\toprule
\multirow{2}*{\textbf{Method}} 
  & \multicolumn{3}{c|}{\textbf{Homophilic Datasets}} 
  & \multicolumn{3}{c}{\textbf{Heterophilic Datasets}}\\
  & Cora & Citeseer & PubMed & Cornell & Wisconsin & Texas \\
\midrule
Homophily & 0.81 & 0.80 & 0.74 & 0.30 & 0.21 & 0.11\\
Nodes     & 2,708 & 3,312 & 19,717 & 183 & 251 & 183\\
Classes   & 7 & 6 & 3 & 5 & 5 & 5\\
\midrule
\modelname\textsubscript{(Heat)}   & 88.16          & 76.92          & 89.73  & 73.52 & 67.46 & 64.41 \\
\modelname\textsubscript{(Wave)}   & \best{89.62}   & \best{77.16}   & 89.63         & 76.44 & 78.47 & 81.24 \\
\modelname\textsubscript{(Sprott)} & 60.81          & 67.42          & 64.18         & \best{88.19} & \best{86.72} & \best{87.63} \\
MPNN                                & 83.93          & 72.81          & 80.43         & 65.17 & 65.29 & 45.87 \\
GAT                                 & 87.28          & 75.03          & 86.94         & 54.27 & 59.14 & 48.62 \\
GIN                                 & 88.95          & 76.04          & \best{89.74}  & 74.68 & 68.47 & 73.87 \\
GWT                                 & 86.23          & 75.92          & 88.37         & 70.34 & 66.25 & 62.11 \\
GraphGPS                            & 87.31          & 75.87          & 88.91         & 73.95 & 69.13 & 74.01 \\
GRAND                               & 84.18          & 73.62          & 80.39         & 81.94 & 74.65 & 77.06 \\
GRAND\texttt{++}                    & 84.33          & 75.61          & 80.53         & 80.27 & 78.38 & 82.58 \\
\bottomrule
\end{tabular*}%
}
\end{table*}

\begin{table*}[!htbp]
\renewcommand{\arraystretch}{1.1}
\small
\caption{MSE (lower is better) for extended persistence image prediction (clustering‐coefficient filtration, degree features).}
\label{tab:appendix_clustering_coeff_epi}
\centering
\begin{tabular}{lccc}
\toprule
\textbf{Model}           & \textbf{Cora}      & \textbf{Citeseer}  & \textbf{PubMed}   \\
\midrule
\modelname{}\textsubscript{(Heat)}      & 2.48                  & 7.35               & \best{1.28}  \\
\modelname{}\textsubscript{(Wave)}      & \best{1.76}           & \best{2.45}        & 6.04         \\
\modelname{}\textsubscript{(Sprott)}    & 8.37                  & 13.58              &  6.26        \\
MPNN                     & 4.20               & 12.6               & 7.94              \\
GAT                      & 9.25               & 4.45               & 7.50              \\
GIN                      & 9.89               & 7.34               & 2.17              \\
GWT                      & 4.12               & 6.94               & 3.22              \\
GraphGPS                 & 14.3               & 11.7               & 9.45              \\
GRAND                    & 13.1               & 10.8               & 6.07              \\
GRAND\texttt{++}         & 15.1               & 6.49               & 4.75              \\
\hline
Neural EPD Approx.       & 9.38               & 1.94        & 4.52              \\
\bottomrule
\end{tabular}
\end{table*}


\section{Dataset Description}
\label{app:datasets}

\textbf{Cora}~\citep{mccallum_automating_2000} is a citation network comprising 2,708 scientific publications classified into one of seven categories. Each node represents a publication and is associated with a 1,433-dimensional binary feature vector indicating the presence or absence of specific words from a predefined dictionary. Edges represent the 5,429 citation links between documents.

\textbf{Citeseer}~\citep{giles_citeseer_1998} is a citation network containing 3,312 scientific publications categorized into six classes. Each node corresponds to a publication and is described by a 3,703-dimensional binary feature vector based on the presence or absence of specific dictionary words. The graph includes 4,732 citation links, forming edges between related documents.

\textbf{PubMed}~\citep{sen_collective_2008} is a citation network of 19,717 biomedical research articles from the PubMed database, all related to diabetes, and categorized into three classes. Each node represents a publication and is associated with a 500-dimensional feature vector based on TF-IDF weighted word frequencies. The graph contains 44,338 citation edges.

\textbf{Cornell}, \textbf{Texas}, and \textbf{Wisconsin}~\citep{pei2020geom} are subgraphs extracted from the WebKB dataset, comprising webpages from the computer science departments of the respective universities. Each node represents a webpage, described by a bag-of-words feature vector derived from its textual content. Edges correspond to hyperlinks between pages. The classification task involves predicting the type of webpage (e.g., student, faculty, course, project, staff). These graphs are relatively small, with 183-251 nodes and 295-499 edges. Notably, all three datasets exhibit strong heterophily, where connected nodes often belong to different classes, posing a challenge for traditional homophily-based graph learning methods.

\textbf{ogbn-papers100M} and \textbf{ogbn-mag} are large-scale academic graphs from the Open Graph Benchmark (OGB) collection. ogbn-papers100M is a directed citation network comprising over 111 million papers indexed in the Microsoft Academic Graph (MAG)~\citep{10.1162/qss_a_00021}, where each node represents a paper with a 128-dimensional word2vec feature vector, and edges denote citation links. ogbn-mag is a heterogeneous graph also derived from MAG, containing four node types - papers (736K), authors (1.1M), institutions (8.7K), and fields of study (60K) - and four directed edge types: authorship, citation, affiliation, and topic assignment. Only paper nodes have input features (128-dimensional word2vec embeddings), while the other node types are featureless.

\textbf{PROTEINS}~\citep{dobson_distinguishing_2003}, part of the TUDataset benchmark suite~\citep{morris2020tudataset}, is a graph classification dataset consisting of 1,113 protein structures, each labeled as either an enzyme or a non-enzyme. In each graph, nodes represent amino acids, and edges are formed between pairs of amino acids that are within 6 Ångströms of each other in 3D space. 

\textbf{ENZYMES}~\citep{schomburg_brenda_2004}, part of the TUDataset benchmark suite~\citep{morris2020tudataset}, contains 600 protein tertiary structures categorized into six enzyme classes, as defined by the BRENDA enzyme database. Each protein is represented as a graph, where nodes correspond to amino acids and edges capture spatial or sequential proximity.

\textbf{MUTAG}~\citep{kriege_subgraph_2012}, part of the TUDataset benchmark suite~\citep{morris2020tudataset}, is a graph classification dataset consisting of 188 chemical compounds labeled according to their mutagenic effect on Salmonella typhimurium. Each compound is represented as a graph, where nodes correspond to atoms (with one-hot encoded atom types as features) and edges represent chemical bonds. There are 7 discrete node labels. The task is to predict the binary mutagenicity label based on molecular structure.

\textbf{DrugBank}~\citep{wishart_drugbank_2018} is a publicly available resource that integrates detailed information about drugs and their molecular targets. We use version 5.0 of the database, released in 2018, which contains 6,712 drug entries, including 1,448 FDA-approved small-molecule drugs. While the database includes a wide range of chemical, pharmacological, and structural properties, we focus on predicting two geometry- and topology-related molecular attributes: total polar surface area (TPSA) and the number of aromatic rings. Each molecule is represented as a graph, with atoms as nodes and bonds as edges.

The \textbf{Materials Project (MP)} dataset~\citep{10.1063/1.4812323} consists of a large collection of inorganic compounds labeled with physical and chemical properties computed using density functional theory (DFT). We use version 2018.6.1, which includes 69,239 materials and a range of properties such as formation energy, bulk and shear moduli, and electronic band gap. In our experiments, we focus on predicting the band gap (e.g., in eV), a key electronic property available for 45,901 compounds. Each material is represented as a graph, with atoms as nodes and edges defined by interatomic bonds or distances derived from crystal structures.

The \textbf{Antiviral Screen Dataset}~\citep{nih_aids_screen} originates from the Drug Therapeutics Program (DTP) AIDS Antiviral Screen, which evaluated the anti-HIV activity of 43,850 chemical compounds based on their ability to inhibit HIV replication. Each compound is represented as a molecular graph, with atoms as nodes and bonds as edges. Screening outcomes were originally categorized into three groups: confirmed active (CA), confirmed moderately active (CM), and confirmed inactive (CI). As part of the MoleculeNet benchmark~\citep{wu_moleculenet_2017}, the CA and CM categories are merged, resulting in a binary classification task: predicting whether a compound is active (CA/CM) or inactive (CI).

\end{document}